\documentclass[final,3p,times,number,sort&compress]{article}
\usepackage[T1]{fontenc}
\usepackage{amsmath,amssymb,amsthm,mathtools}
\usepackage{booktabs,tabularx,longtable,array}
\usepackage{microtype}
\usepackage{placeins}
\usepackage[skins]{tcolorbox}
\usepackage[hidelinks]{hyperref}
\usepackage{xurl}
\usepackage[margin=2.5cm]{geometry}
\newcommand{\E}{\mathbb E}
\newcommand{\KL}{D_{\mathrm{KL}}}
\newcommand{\CE}{\operatorname{CE}}
\newcommand{\Gain}{\operatorname{Gain}}
\newcommand{\cX}{\mathcal X}
\newcommand{\cY}{\mathcal Y}
\newcommand{\cC}{\mathcal C}
\newcommand{\cM}{\mathcal M}
\newcommand{\cZ}{\mathcal Z}
\newcommand{\vis}{\operatorname{vis}}
\newtheorem{proposition}{Proposition}

\newtcolorbox{runningexample}[1]{
  enhanced,colback=black!2,colframe=black!45,
  colbacktitle=black!8,coltitle=black,fonttitle=\bfseries,
  boxrule=0.4pt,arc=0pt,left=6pt,right=6pt,top=5pt,bottom=5pt,
  before skip=0pt,after skip=0pt,
  title={#1}}
\title{From Human Corrections to Learned Segmentations:\\
A Critical Mathematical Review of\\ Information, Learning, and Human Effort}
\author{Jon Sporring\\Department of Computer Science,\\University of Copenhagen,\\DK-2100 Copenhagen,\\Denmark\\e-mail:
  sporring@di.ku.dk}
\begin{document}
\maketitle
\begin{abstract}
When does a human correction become useful learning, and what establishes
better segmentation at matched human effort? This critical narrative review
follows feedback through its selection, learner update, and evaluation. It
compares active learning, machine teaching, interactive segmentation, and
sequential predictive evaluation using a common mathematical account of
target information, predictive error, response predictability, and human work.
An exact Bayesian update converts evaluation-label information into expected
log-loss improvement; an excess-risk decomposition explains what an
approximate learner can lose. Conditional information decompositions separate
the roles of selected locations, labels, and displayed context. Empirical
comparisons show why response protocols, update mechanisms, and effort
denominators matter, distinguishing current-mask refinement from reusable
learning and simulated interaction from human evidence. We identify what
logs can reveal, what requires independent evaluation, and what remains
model-dependent. The resulting research priorities are to test what selection
reveals, whether using it improves assimilation, and whether the resulting
interaction policy benefits segmentation at matched effort. These are distinct
claims: predicting a teacher's responses need not identify human intentions
or establish better learning. No new acquisition algorithm or empirical
superiority is claimed.\\
\textbf{keywords:} Interactive segmentation, Machine teaching, Active learning,
Human feedback, Information gain, Minimum description length
\end{abstract}

\paragraph{Reading guide}
The review follows a correction from selection to learner update and
evaluation. A target specifies the desired label distributions; a response
records what the human chooses to mark; a predictor determines the resulting
segmentation error. These are distinct objects. Table~\ref{tab:notation}
provides a quick reference, and boxed figures illustrate the distinctions
on four pixels, beginning with Figure~\ref{fig:example-setup}. The literature selection is described in
Section~\ref{sec:review-method}. Standard derivations are in
\ref{app:derivations}; the qualified parametric MDL benchmark is in
\ref{app:mdl}. All logarithms are base two unless stated otherwise.

\clearpage
\section{When do corrections become useful learning?}
\label{sec:question}
Corrective segmentation combines a prediction, a human judgment of that
prediction, and an update. The same mark may refine the current mask, train a
model for later views, or do both. RootPainter exemplifies the latter setting:
corrections accumulate while a task-specific segmentation model is trained
\cite{smith2022rootpainter,smith2022rootpainter3d}. The central question is not
merely how many labels have been collected, but what follows from observing
them, updating a learner, and comparing the resulting outcomes:

\begin{quote}
When is corrective feedback informative, when does it become useful learning,
and what does a claimed improvement per unit of human effort establish?
\end{quote}

Existing approaches answer different versions of this question. Query
selection, pedagogical example selection, mask refinement, and sequential
compression differ both in the decisions they control and in what counts as
success. A common mathematical description lets us compare these assumptions
without treating every method as an approximation to a single objective.

We use \emph{progressive teaching} for the coupled corrective process, not for
a competing acquisition algorithm or a claim that humans teach optimally.
The review is organised around three questions:
\begin{enumerate}
 \item Which participant selects each part of the supervision, and what
       observation model makes that selection informative?
 \item Does a method seek target information, predictive improvement, or
       economical use of human work, and when do these objectives agree?
 \item Which assumptions connect acquired corrections to lasting learning,
       and what evidence is needed when those assumptions fail?
\end{enumerate}
The contribution is a critical mathematical comparison of the assumptions
connecting these three questions. We distinguish the evidence in a selected
response from the effect of a learner update and from the resource comparison
needed to call that effect efficient. These distinctions concern both current-
image refinement and reusable predictors; success at one does not establish
success at the other. RootPainter motivates the reusable-learning case, but
does not define the review's conclusion. A teacher-response model is useful
only if it helps explain or improve a specified outcome; predicting familiar
strokes is not itself evidence of better learning.

\section*{Notation at a glance}
Random objects and their realized values are distinguished explicitly where
needed, for example $R_i$ and $r_i$, or $M_i$ and $\cM_i$. All information
quantities use base-two logarithms. The table previews notation defined in
the main text; the boxed figures develop one four-pixel example.

\begingroup
\renewcommand{\arraystretch}{1.12}
\begin{longtable}{@{}p{0.24\linewidth}p{0.70\linewidth}@{}}
\caption{Core symbols, meanings, and relations.}\label{tab:notation}\\
\toprule
Symbol & Meaning and principal relation\\
\midrule
\endfirsthead
\caption[]{Core symbols, meanings, and relations (continued).}\\
\toprule
Symbol & Meaning and principal relation\\
\midrule
\endhead
\midrule
\multicolumn{2}{r}{Continued on next page}\\
\endfoot
\bottomrule
\endlastfoot
\multicolumn{2}{@{}l}{\textit{Images, targets, and predictions}}\\*
$J,\cX,\cY$ & Fixed image collection, address domain, and label alphabet; $J:\cX\to\mathbb R^d$.\\
$X\sim\nu$ & Evaluation address and weights; $\nu$ is not the view-selection distribution.\\
$F\sim\mu$, $f$ & Random interpretation across sessions and its fixed session realization.\\
$P_F$, $P_f$ & Random target label distribution and its realization; $P_f:\cX\to\mathcal P(\cY)$.\\
$T_F$, $T_f$ & Random and realized evaluation-equivalence classes; members of $T_f$ agree with $P_f$ $\nu$-almost everywhere.\\
$S_{i-1},Q_i,\mathcal Q$ & Pre-response state, display predictor $Q_i=Q(\cdot\mid\cdot,S_{i-1})$, and learner family.\\
\multicolumn{2}{@{}l}{\textit{Interaction and memory}}\\*
$\mathcal X_i,\mathcal V_i$ & Realized view and display; marked addresses need not fill the view.\\
$Z_i$, $\cZ_i$ & Random pre-response memory and its realization $\cZ_i=(\cM_{i-1},\mathcal X_i,\mathcal V_i)$.\\
$R_i=(\Xi_i,H_i)$ & Random response: marked support and label assignment.\\
$r_i=(\xi_i,h_i)$ & Realized response; $\xi_i\subseteq\mathcal X_i$, $h_i:\xi_i\to\cY$.\\
$M_i$, $\cM_i$ & Random post-response memory $M_i=(Z_i,R_i)$ and realization $\cM_i=(\cZ_i,r_i)$.\\
$\Lambda,\vis,\Phi$ & View selector, visualization function, and learner state update.\\
$\rho_f$ & Teacher-response distribution at fixed interpretation: $R_i\sim\rho_f(\cdot\mid Z_i=\cZ_i)$.\\
$\rho_i$, $\widehat\rho_i$ & Source-marginal response distribution given context, and shared causal coding model.\\
\multicolumn{2}{@{}l}{\textit{Beliefs, information, and effort}}\\*
$\cC_i$, $w(P\mid\cM)$ & Toy candidate set and learner model weights; neither is the target class or necessarily the source posterior.\\
$\widetilde Q$, $\widehat Q$ & Uniform toy candidate-average predictor and posterior mixture used in BALD.\\
$\CE(P_f,Q)$, $\tau(T_f,Q)$ & Cross-entropy risk and evaluation-weighted KL distortion; $\CE=H_f(Y\mid X)+\tau$.\\
$\Gain_i$ & Full-round target-entropy reduction, including any new context information; not necessarily predictive improvement.\\
$W_n$ & Random marked-location work $W_n=\sum_{i=1}^n|\Xi_i|$; repeats count, elapsed time is not implied.\\
$C_n$ & Measured human cost $\sum_{i=1}^n c_i$ in a declared unit; $C_n=W_n$ is the mark-count specialization.\\
$\ell_i$, $\ell_{\mathrm{PT}}(n)$ & Ideal response length $\ell_i=-\log_2\widehat\rho_i(R_i\mid Z_i)$ and cumulative expectation $\ell_{\mathrm{PT}}(n)=\E\sum_{i=1}^n\ell_i$.\\
\end{longtable}
\endgroup

\section{Related perspectives: who, how, and what makes it efficient?}
\label{sec:literature}
\subsection{Review questions and literature selection}
\label{sec:review-method}
This is a critical narrative review, not a systematic survey or meta-analysis.
We select work by its observation model, learner update, or evaluation design,
not by whether it favours teacher modelling. Foundational methods explain the
criteria; primary segmentation studies supply empirical evidence. Architecture
variants without a distinct interaction or evaluation mechanism are excluded.

The detailed studies in Table~\ref{tab:interaction-evidence} span reusable
learning (Crayons, RootPainter, RootPainter3D, Cellpose, XpertLoop), image-specific
adaptation (BIFSeg), prompting and reuse (microscopy SAM), workflow integration
(MONAI Label), response modelling (RClicks), and cost-aware acquisition
(CEREALS). This contrast-based selection is not a representative statistical
sample. Adjacent studies test other feedback units, contextual reuse, and
cross-image adaptation; they inform scope rather than the table's estimates.

Prior reviews provide candidate references. A bounded primary-source search
through 14 October 2026 and targeted citation tracing supplement them for
corrective segmentation, response modelling, and reusable learning.
\ref{app:search-record} records queries, inspected versions, and
selection decisions. Coverage is not exhaustive or independently
double-screened. Preprints are identified explicitly.

Comparisons illustrate mechanisms, not maximal effects.
\ref{app:evidence-appraisal} records population and horizon, intervention
and controls, endpoint and effort, uncertainty unit, selection rationale, and
qualifying alternative results. Missing reporting is distinguished from
design limitations. Heterogeneous outcomes are not pooled. Equations give
qualified specializations; examples establish possibilities, not rankings.

\subsection{Relation to existing reviews}
Dudley and Kristensson organise interactive machine learning around the user,
model, data, and interface, including incomplete feedback and the difficulty
of translating user intent into input \cite{dudley2018review}. Wu et al.
review human involvement in data processing, model training, and system
design \cite{wu2022human}; Ren et al. survey deep active-learning strategies
and their practical challenges \cite{ren2021survey}. For medical image
segmentation, Marinov et al. distinguish interactive learning, interactive
inference, and interactive online learning, and examine simulated-user realism,
missing baselines, and inadequate user-centred evaluation
\cite{marinov2024review}. These reviews establish both the human--learner
loop and the need to scrutinize its evaluation.

Our narrower comparison asks which components of a spatial response are
selected by whom, what observation model makes that selection informative,
and whether an improvement concerns the current mask or a reusable learner.
It connects these distinctions to target information, predictive distortion,
response predictability, and human work. The additional contribution is a
formal account of the assumptions connecting observation, update, and effort:
two methods with similar interfaces can optimize
different quantities, while methods from different applications can share
an observation or cost assumption. The common notation is useful only insofar
as it exposes these differences and their consequences for evaluation.

Four transfer tests organise the synthesis. Does a requested-label model
describe a human-selected support? Does better current-mask quality show
learning for later images? Does a more predictable response indicate more
useful supervision? Does a smaller mark count establish less human effort?
The answer to each depends on an explicit observation, update, or measurement
assumption, rather than on the method's name. Sections~\ref{sec:gain},
\ref{sec:work}, and~\ref{sec:coding} make those assumptions mathematical;
Section~\ref{sec:research} turns them into criteria for interpreting evidence.

\subsection{Who selects supervision, and how do the participants interact?}
In active learning, the learner selects a query. In machine teaching, the
teacher selects supervision with a learner in mind. Corrective segmentation
can combine both: the system proposes a view, the teacher chooses where to
mark, and either participant may initiate navigation. This allocation of
choices is more informative than classifying the entire application as only
active learning or only teaching \cite{settles2009active,zhu2018overview}.

Iterative machine teaching explicitly adapts examples to learner state
\cite{liu2017iterative}; cross-space work relaxes shared-representation
assumptions \cite{liu2018blackbox}. Interactive machine learning studies the
human--learner loop more generally \cite{amershi2014power}, while coactive
learning studies feedback that improves a structured prediction without
requiring an optimal correction \cite{shivaswamy2015coactive}. Its regret
analysis assumes informative feedback, with variants allowing slack or
informativeness in expectation; an arbitrary correction need not satisfy
these assumptions. Communicative
learning places mutual modelling within a broader account of learning as
information exchange \cite{yuan2023communicative}.

Pedagogical reasoning makes the selection mechanism explicit: Shafto et al.
model teachers choosing examples for a learner and learners interpreting that
choice \cite{shafto2014rational}. Their concept-learning examples include
both point locations and labels. Iterative teacher-aware learning transfers
this idea to parameter learning through a model of why a teacher selected
an example \cite{yuan2021teacheraware}. Its update compares the chosen
example with labelled alternatives under a heuristic teacher-selection
model. Such alternatives are generally unlabelled in sparse corrective
segmentation. Moreover, a mismatch between assumed and actual teaching
behaviour can remove the benefit; teacher-awareness is not automatically
better than treating examples as ordinary samples.

Consequently, neither having a history nor inferring information from a
teacher's choice is a novelty claim by itself. The application-specific
question is what is gained by jointly representing displayed predictions,
selected spatial supports, label values, learner updates, and human work.

\subsection{What changes when a segmentation is corrected?}
Classical interactive segmentation uses sparse input to constrain an image
segmentation objective \cite{boykov2001interactive,rother2004grabcut}. Modern
refinement systems often evaluate quality after a given number of clicks,
or clicks required to reach a threshold
\cite{benenson2019large,sofiiuk2022focalclick,liu2023simpleclick}. This differs
from training a reusable pixel classifier: already in Crayons, users painted
examples, inspected classifications, and corrected errors to train a model
\cite{fails2003interactive}. Trainable Weka Segmentation and ilastik likewise
learn classifiers from interactive annotations for application to further
data \cite{arganda2017weka,berg2019ilastik}. RootPainter belongs to this
established corrective-learning lineage, with a deep learner and sparse
corrective supervision \cite{smith2022rootpainter,smith2022rootpainter3d}.

There are intermediate and hybrid regimes. BIFSeg adapts network parameters
to the particular image being segmented, using a bounding box and scribbles;
updating parameters therefore need not imply learning for later images
\cite{wang2018bifseg}. Cellpose 2.0 alternates correcting predicted masks,
retraining, and processing subsequent images \cite{pachitariu2022cellpose}.
Its corrected masks provide different supervision from sparse labels whose
unmarked surroundings remain unknown. MONAI Label integrates interactive
annotation, model training, and active-learning services
\cite{diazpinto2024monai}; supporting these components is distinct from
establishing the benefit of a particular acquisition policy.

Promptable models add another distinction. Segment Anything conditions a
pretrained model on spatial prompts to produce a mask
\cite{kirillov2023sam}. Accumulating prompts can improve that mask without
updating the model parameters. Segment Anything for Microscopy combines
interactive annotation with microscopy-specific models and fine-tuning,
and evaluates annotation time and quality with users
\cite{archit2025microscopy}. Prompt state, image-specific adaptation, and
reusable-model training should therefore be recorded separately rather than
identified by the presence of an interactive interface.

Reuse need not change weights: MultiverSeg retains completed image--mask pairs
as context, whereas \emph{You Point, I Learn} updates parameters during and
after interaction \cite{wong2025multiverseg,xu2026youpoint}. The HiTTA preprint
also carries corrected-mask updates forward \cite{hu2024hitta}. All three
evaluate simulated interactions; HiTTA uses existing annotators' masks to
simulate preferences and evaluates the better of two prediction heads.

Live studies address different endpoints. MedSAM times initialization,
inference, and refinement for two radiologists completing ten adrenal-tumour
cases, with fixed task order \cite{ma2024medsam}; it does not
test reusable learning. SPA adapts a latent distribution from preferences
among candidate masks and tests cross-image transfer of that state; its
five-participant study reports shorter interactions than MedSAM
\cite{zhu2025spa}. Preference-state reuse differs from modelling natural
spatial-support selection.

Corrections and reuse coexist in two preprints: Dendukuri et al.'s timed
surgical-video study, where SAM2 assistance outpaces iterative training
\cite{dendukuri2026active}, and Shaeri et al.'s critic interface with
fine-tuning and cross-image propagation \cite{shaeri2025explainable}.
ExiL combines simulated refinement trajectories, expert-approved updates,
and single-expert timing \cite{tavangar2026exil}. These precedents establish
history-based correction without estimating natural human selection probabilities.

\paragraph{Human responses versus simulated clicks}
The response protocol specifies which errors are noticed, corrected, or
ignored. RClicks collects real click placements and learns a clickability
model for more realistic segmentation benchmarking
\cite{antonov2024rclicks}. Its data include later-click tasks constructed
from selected model-error masks; these are not unrestricted longitudinal
teaching sessions. It establishes a direct precedent for modelling spatial
response support, but does not establish that using such a model improves
a reusable learner. This separates three claims: realistic response
prediction, robust mask refinement, and more effective corrective training.
SegAgent provides a complementary caution: its human-like annotation
trajectories are generated from reference-mask errors, not recorded human
teaching sessions \cite{zhu2025segagent}. Imitation of a click generator does
not by itself validate the generator as a human-response model.

\subsection{What does the empirical evidence establish?}
\label{sec:study-evidence}
Table~\ref{tab:interaction-evidence} compares findings within their protocols,
not scores across tasks. SD denotes standard deviation and SE standard error;
neither is a confidence interval. Counts and figure/table locators refer to
the cited experiments.
Dice, downstream measurement correlations, and mean segmentation accuracy
(mSA) remain separate endpoints: none is converted to KL distortion.

\begingroup
\small
\setlength{\tabcolsep}{4pt}
\begin{longtable}{@{}>{\raggedright\arraybackslash}p{0.2\linewidth}>{\raggedright\arraybackslash}p{0.34\linewidth}>{\raggedright\arraybackslash}p{\dimexpr0.45\linewidth-4\tabcolsep\relax}@{}}
\caption{Study-level evidence: selection, assimilation, and human effort.
Prompt refinement, current-image adaptation, and reusable-model training are
distinct update scopes. ``Time'' and counts are not interchangeable effort
measures.}\label{tab:interaction-evidence}\\
\toprule
Study & Selection, feedback, and update & Finding, uncertainty, and scope\\
\midrule
\endfirsthead
\multicolumn{3}{l}{\tablename~\thetable\ (continued)}\\
\toprule
Study & Selection, feedback, and update & Finding, uncertainty, and scope\\
\midrule
\endhead
\midrule
\multicolumn{3}{r}{Continued on next page}\\
\endfoot
\bottomrule
\endlastfoot
Crayons \cite{fails2003interactive}
& Human-painted examples and corrections train a reusable pixel classifier;
  the comparison changes the fitting algorithm, not a query policy.
& Ten users, counterbalanced order: skin-classification time to 97.5\%
  reference agreement averaged 3:06/3:51 minutes for the stronger-generalizing
  algorithm versus 10:35/10:26 for the faster-fitting alternative in the two
  order groups (Tables 1--2). No intervals or tests are reported. Slower
  individual updates can therefore accompany faster interactive completion.\\[5pt]
RootPainter \cite{smith2022rootpainter}
& Humans select foreground/background marks after viewing predictions;
  sparse corrections accumulate in reusable-model training. Initial clear
  examples precede corrective annotation.
& Two author-annotators, three datasets: time-matched corrective versus dense
  annotation gave held-out measurement $R^2$ of 0.80 (SE 0.04) versus
  0.75 (SE 0.07), with no significant strategy effect at the 0.05 threshold
  (Table 3). Corrective annotation came first. The comparison establishes
  neither superiority nor equivalence.\\[5pt]
RootPainter3D \cite{smith2022rootpainter3d}
& A physician selects spatial corrections; a model is trained throughout
  annotation of successive CT volumes. Both completed contours and the
  reusable predictor change.
& One physician, 933 hearts: the last ten averaged 122 seconds with assistance
  versus 421 manually (Fig. 5); no timing interval is reported. In 20 paired
  contour comparisons, corrected-output Dice was 0.952 (SD 0.007) against
  manual contours (Table 1). Timings include inspection but exclude inactivity
  of at least 20 seconds; the design does not isolate selection modelling.\\[5pt]
BIFSeg \cite{wang2018bifseg}
& Human bounding boxes and scribbles guide segmentation and image-specific
  network adaptation. Parameters change, but the intervention is targeted
  to the current image.
& Two clinical users. With identical initial predictions and one-round
  scribbles, tumour-core Dice was 87.49\% versus 85.93\% for CRF, and
  whole-tumour Dice 88.11\% versus 85.18\% ($p<0.05$, paired tests;
  Table VI). This is evidence of better use of the same corrections for
  current-case refinement, not accumulated learning across cases.\\[5pt]
Cellpose 2.0 \cite{pachitariu2022cellpose}
& A human corrects and completes instance masks; accumulated masks retrain a
  reusable model before the next image. The design compares online/offline
  annotation and pretrained/from-scratch initialization.
& One annotator, two categories: online/pretrained annotation used 167 versus
  663 and 158 versus 897 manually drawn objects compared with offline
  annotation (Fig. 5). Test performance approached within-annotator agreement
  with 100--200 objects; cost-ratio intervals are not reported. Pretraining
  and repeated fitting contribute jointly; object counts are not session time.\\[5pt]
MONAI Label \cite{diazpinto2024monai}
& The framework supports system query selection and human clicks/scribbles,
  with mask refinement and reusable-model retraining. The reported workflow
  is not one fixed acquisition method.
& DeepGrow validation Dice rose from 0.891 to 0.967 (2D) and 0.730 to 0.959
  (3D) across annotation stages (Table 2), without uncertainty intervals.
  These are workflow demonstrations, not controlled acquisition-policy
  comparisons. Inconsistent validation counts and timing totals in the
  source prevent a precise cost-effect estimate.\\[5pt]
RClicks \cite{antonov2024rclicks}
& Humans select click locations for specified targets/error displays. A
  learned response-location model supplies simulated clicks for evaluating
  current-mask refinement, not within-session reusable-model training.
& 475{,}544 real clicks support the response model. Its multi-click simulations
  require approximately 5--29\% more clicks than conventional simulated
  clicking to reach 90\% IoU, capped at 20 clicks (Table 4).
  Reported SD measures click-group variability, not uncertainty in measured
  human time. This tests benchmarking sensitivity, not reusable teaching.\\[5pt]
Microscopy SAM \cite{archit2025microscopy}
& Human prompts refine masks; dataset-specific fine-tuning is evaluated
  separately from interactive prompting. Automatic predictions can then
  serve further annotation.
& Five organoid annotators: fine-tuned $\mu$SAM and CellPose both averaged
  0.7 seconds/object. Annotation mSA was 0.72 versus 0.76, but test mSA was
  $0.51\pm0.02$ versus $0.42\pm0.05$ (SD across annotators; Fig. 6a).
  Point estimates for completion and later generalization rank differently; rates do
  not measure complete training-plus-human cost.\\[5pt]
CEREALS \cite{mackowiak2018cereals}
& The learner chooses image regions using uncertainty and predicted
  annotation cost; an oracle reveals existing labels. Segmentation and cost
  models are refitted after acquisition.
& Five simulated Cityscapes runs: cost-aware selection reached 95\% of
  full-data mIoU at 17.07\% of recorded polygon-click cost, versus 33.76\%
  for entropy-only selection of $128\times128$ regions (Figs. 4--5).
  Threshold-cost intervals are not reported. This supports cost-aware
  acquisition under the robot oracle, not a measured human-time reduction.\\[5pt]
XpertLoop \cite{jaiswal2026xpertloop}
& Experts correct presegmentations for iterative retraining; random and
  active case selection are compared. Timing compares manual annotation
  with correction of optimized-model predictions.
& One annotator per timed task: maximum time savings were 90.3\% for kidneys
  and 48.2\% for tumours ($p<0.001$ and $p=0.003$; Fig. 5).
  Estimated tumour cost savings were \$5.63/examination (95\% CI
  $-7.26$ to $26.09$), using wage/time modelling rather than user-population
  uncertainty. This is not a matched end-to-end training-budget comparison.\\
\end{longtable}
\endgroup

Three links emerge. \emph{Assimilation:} BIFSeg holds corrections fixed while
changing their use; Crayons trades fitting speed against generalization.
These change the update $\Phi$, not merely the information received.
Equation~\eqref{eq:assimilation-gap} provides a log-loss account of this
distinction, not an information-theoretic reinterpretation of their Dice
or completion-time effects.

\emph{Selection:} RClicks tests a human-data-based response simulator, whereas
CEREALS changes acquisition $\Lambda$ under a specified cost proxy. Neither tests
the complete chain from predicting natural support choices to better
reusable learning. Transferring simulated gains requires a representative
response model and a validated effort denominator.

\emph{Efficiency:} RootPainter's matched-time estimate is inconclusive;
RootPainter3D and XpertLoop show completion savings under specific training
histories. Cellpose's pretraining comparison and microscopy SAM's different
completion/test rankings show why initialization, update horizon, and outcome
must be held explicit. Late or optimized-model savings are not whole-session
savings. These distinctions transfer across losses; the KL identities do not
convert the studies' endpoints into distortion.

\subsection{Which labels should the learner request?}
For exposition, let a learner assign weights $w(P\mid\cM)$ to a finite collection
$\cC$ of candidate label distributions after history $\cM$. Its posterior
predictive distribution is
\begin{equation}
 \widehat Q(y\mid x,\cM)=\sum_{P\in\cC}w(P\mid\cM)P(y\mid x),
 \qquad w\geq0,\quad \sum_Pw(P\mid\cM)=1.
 \label{eq:bald-mixture}
\end{equation}
Sums become integrals for continuous model posteriors. This is an explicit
modelling choice, not an enumeration of every possible segmentation function.
Approximate posteriors or ensembles may make the calculation tractable but can
miss plausible targets.

\paragraph{Predictive entropy and BALD}
Predictive-entropy acquisition ranks $x$ by $H(\widehat Q(\cdot\mid x,\cM))$.
BALD instead uses the label's mutual information with the candidate model
\cite{houlsby2011bald,houlsby2012collaborative}:
\begin{equation}
 \operatorname{BALD}(x)=H(\widehat Q(\cdot\mid x,\cM))
 -\sum_Pw(P\mid\cM)H(P(\cdot\mid x)).
 \label{eq:bald}
\end{equation}
It subtracts label variability shared by individual candidate models. If every
candidate predicts $(1/2,1/2)$, predictive entropy is one bit but BALD is zero.
If all candidates are deterministic, their entropies vanish and the criteria
coincide. BALD uses a posterior predictive distribution for acquisition; it need
not prescribe the application's deployed predictor. The distinction between
model uncertainty and within-model label variability is also important in
segmentation uncertainty modelling \cite{kendall2017uncertainties}.

\paragraph{BatchBALD and diversity}
For a batch $B$, write $P_B$ for a candidate's joint label distribution and
$\widehat Q_B=\sum_Pw(P\mid\cM)P_B$. BatchBALD uses
\begin{equation}
 \operatorname{BatchBALD}(B)=H(\widehat Q_B)-\sum_Pw(P\mid\cM)H(P_B).
 \label{eq:batchbald}
\end{equation}
Joint scoring accounts for redundant information that a sum of individual
scores misses \cite{kirsch2019batchbald}. The published construction uses
conditional independence of labels given model parameters and greedy batch
selection. Marginal label distributions alone do not determine $P_B$ without
such an assumption or an explicit joint model. Nor does scoring all pixels in
a view model which subset a teacher will actually annotate. Core-set
acquisition instead seeks geometric batch coverage in a learned
representation \cite{sener2018active}. Its image-classification evidence
makes it a useful diversity baseline, not a guarantee of coverage of rare
segmentation errors.

\paragraph{Expected improvement and variance reduction}
Cohn et al. analyse squared prediction error, use a negligible-bias assumption,
and acquire labels to reduce expected prediction variance
\cite{cohn1996active}. In a discrete evaluation domain, their idea can be
written schematically as
\begin{equation}
 x_{\mathrm{next}}\in\arg\min_x
 \E_{Y\sim\widehat Q(\cdot\mid x,\cM)}
 \left[\sum_z\nu(z)v_{x,Y}(z)\right],
 \label{eq:variance}
\end{equation}
where $v_{x,Y}(z)$ is the estimated variance after fitting the additional
example. This is a precedent for selecting supervision by its effect on the
learner, but is not a general guarantee of reduced KL distortion. Related
segmentation methods select for expected change or spatial uncertainty
\cite{vezhnevets2012expected,top2011active}.

\subsection{Who should annotate, and at what cost?}
Yan et al. jointly consider uncertain instances and input-dependent annotator
reliability \cite{yan2011crowds}. Their selection objective can be expressed,
with renamed coefficients, as
\begin{equation}
 \min_{x,\lambda}\left\{
 \eta\left[\operatorname{logit}\widehat Q(1\mid x)\right]^2
 +\sum_a\lambda_a e_a(x)\right\},\qquad
 \lambda_a\geq0,\quad \sum_a\lambda_a=1.
 \label{eq:annotator}
\end{equation}
Here $x$ is their feature-space query, $e_a$ is an annotator noise score, and
$\eta$ balances the terms; the continuous proposal is mapped to the available
pool. The natural-log logit convention can be absorbed into the coefficient
if another base is used. This is an annotator-selection model, not a model of
human-selected spatial correction support.

CEREALS combines informative regions with estimated annotation cost in
semantic segmentation \cite{mackowiak2018cereals}. Haertel et al. study
return-on-investment acquisition and conditions relating local benefit/cost
selection to performance across a budget, with experiments in part-of-speech
annotation \cite{haertel2015roi}. Its relevance is the cost--benefit structure,
not a claim about pixelwise KL per mark. A ratio criterion is not in general
an optimal multi-step policy.

XpertLoop combines expert correction timing with reusable-model training and
acquisition comparisons \cite{jaiswal2026xpertloop}. In its reported organ
tasks, size--entropy selection reaches strong performance in fewer loops than
random selection. Plain entropy can underperform random selection, and
suitability varies by dataset. Size--entropy adds inference over unlabelled
cases, while stopping uses test-set performance; loop counts therefore
establish neither equal total costs nor independently assessed gains.

Benenson et al.'s human annotation study makes work-proxy limitations concrete
\cite[Sections~4--5]{benenson2019large}. Under a three-round protocol allowing
up to four clicks per round, zero-click judgments took about four seconds,
and four-click rounds were faster than three-click rounds. Later rounds took
longer, consistent with smaller errors becoming harder to find; a ranker using
interaction-history features predicted completed-mask quality. These findings
motivate accounting for inspection and changing correction behaviour
(Sections~\ref{sec:work} and~\ref{sec:research}), without establishing universal
teaching stages or improved reusable learning. The headline polygon comparison
uses an estimated historical timing denominator, not a matched-user trial.
Alongside Cellpose's correction counts and microscopy SAM's user timings
\cite{pachitariu2022cellpose,archit2025microscopy}, this shows why effort units
and available actions must be specified.

\subsection{What else must a correction model distinguish?}
Probabilistic segmentation can represent multiple masks and spatially correlated
uncertainty \cite{kohl2018probabilistic,monteiro2020stochastic}. These models
represent distributions over complete label maps, not just their pixel
marginals. Zhang et al. jointly estimate a latent consensus-label distribution
and image-/pixel-dependent annotator confusion matrices under explicit
conditional-independence and regularization assumptions
\cite{zhang2023multiple}. This is not an assumption-free separation of valid
alternative interpretations from annotation errors, nor a model of which
regions annotators choose to mark. Error-tolerant segmentation separately
addresses incorrect annotations \cite{wang2020errortolerant}. These
distinctions matter even when the eventual annotation is one hard label.
Human feedback may also depend on the learner's behaviour
\cite{macglashan2017interactive}.

Selective-labels research cautions that the cases on which labels are observed
need not represent the evaluation population \cite{lakkaraju2017selective}.
For correction logs, exposure, attention, error detection, and marking are
distinct filters. Rubin's missing-data analysis gives conditions under which
an observation mechanism can be ignored for inference
\cite{rubin1976missing}. Positive--unlabelled learning distinguishes
selection assumptions and the information needed to identify class
probabilities \cite{bekker2020positive}. The closest corrective analogy is
a detected-and-marked error versus an unmarked location, not foreground
versus background. Confirmatory or erroneous marks require a richer model.

In active learning, Farquhar et al. distinguish acquisition-induced estimation
bias from overfitting and show that removing the former need not improve
deep-model training \cite{farquhar2021bias}. Their estimators use the
acquisition probabilities; displaying an image does not reveal the human's
attention or marking probabilities. These connections explain why
non-correction is not automatically acceptance and why an adaptive log does
not by itself supply an unbiased evaluation sample.

Finally, MDL and prequential coding connect prediction with description length
\cite{rissanen1978modeling,grunwald2007mdl,grunwald2020revisited}. Neural
prequential coding already evaluates how a learning procedure predicts later
labels from earlier ones \cite{blier2018description}. Dawid's prequential
perspective evaluates probability forecasts against subsequently observed
outcomes \cite{dawid1984prequential}; proper scoring rules formalize the
quality of such forecasts \cite{gneiting2007proper}.
Sections~\ref{sec:mdl}
and~\ref{sec:coding} ask what changes when the observations are adaptive spatial
responses rather than a predetermined sequence of labels.

\subsection{Teaching order and adaptation on both sides}
Curriculum learning studies how the selection and order of training examples
affect learning \cite{bengio2009curriculum}. Torpey et al. show that image
order changes Dice and simulated correction counts when completed predictions
become context for later images \cite{torpey2026learning}. Across starting
images, uncertainty-based ordering does not consistently beat random ordering;
its best-start advantage uses retrospective selection. Order sensitivity
therefore does not by itself identify an effective selection policy or
establish human-time savings.
Continual-learning research
organises mechanisms for retaining previous learning, including replay
\cite{delange2022continual}. Its changing-task classification benchmarks
are not the same as a fixed-target corrective session, but they identify
learner-side mechanisms that can make identical supervision produce
different outcomes when presented in different orders.

Human teaching is not necessarily a curriculum optimized for the learner.
Cakmak and Thomaz study natural teaching and learner-specific guidance
\cite{cakmak2014eliciting}; Fiebrink et al. observe users modifying training
strategies and sometimes task goals after evaluating model behaviour
\cite{fiebrink2011evaluation}. The latter study concerns interactive
gesture/music systems rather than segmentation. Together these works
motivate modelling strategy and adaptation, but do not establish a universal
sequence of teaching stages or validate a particular RootPainter protocol.

\subsection{Synthesis: which part of the corrective loop is modelled?}
Table~\ref{tab:mechanisms} compares mechanisms, not benchmark rankings. A
method can address several rows; for example, a cost model can be combined
with a disagreement-based query rule. The key distinction is which choices
are treated as controlled actions, which as observations, and which are
left outside the model.

\begin{table}[htbp]
\centering
\caption{Representative approaches and the assumptions needed to transfer
their interpretation to corrective learning.}
\label{tab:mechanisms}
\begin{tabularx}{\linewidth}{@{}>{\raggedright\arraybackslash}p{0.28\linewidth}>{\raggedright\arraybackslash}X>{\raggedright\arraybackslash}X@{}}
\toprule
Approach & Decision and observation & Issue in the corrective loop\\
\midrule
BALD \cite{houlsby2011bald} & Learner requests a label; beliefs determine its expected informativeness. & A teacher may choose a different support within the displayed view.\\[3pt]
BatchBALD \cite{kirsch2019batchbald} & Learner requests a batch; joint label information accounts for redundancy. & A jointly informative batch is not necessarily the subset the human will label.\\[3pt]
Variance reduction \cite{cohn1996active} & Query scored through its expected effect on a fitted predictor. & Improvement depends on the loss, model assumptions, and actual update.\\[3pt]
Annotator-aware AL \cite{yan2011crowds} & Learner selects an instance and an annotator using reliability estimates. & Reliability and the human's choice of what to correct are distinct mechanisms.\\[3pt]
Iterative teaching and teacher-aware learning \cite{liu2017iterative,yuan2021teacheraware} & Teacher selects supervision; learner may interpret its selection mechanism. & Spatial support, displayed prediction, and human effort require an explicit specialization.\\[3pt]
Interactive segmentation \cite{sofiiuk2022focalclick,smith2022rootpainter} & Human or simulated teacher supplies spatial feedback to a mask or reusable model. & Per-image refinement and learning for subsequent views require different evaluation.\\[3pt]
Prequential evaluation \cite{blier2018description} & A learner predicts successive labels for coding. & An adaptive correction stream includes support selection, not only label values.\\
\bottomrule
\end{tabularx}
\end{table}

This comparison motivates three distinctions developed next: a target is not
an individual response; information in a response is not the improvement
produced by a learner; and predictable supervision is not necessarily
efficient teaching. The formalism below supplies a shared language for these
distinctions, rather than replacing the surveyed algorithms with one new
method. Section~\ref{sec:research} returns to the literature through an
objective comparison and the open problems exposed by these assumptions.

\section{Targets, predictions, and the corrective interaction}
\label{sec:framework}
\subsection{A fixed collection and a distribution-valued target}
Fix an image collection $J:\cX\to\mathbb R^d$, a finite address domain
$\cX$, and a finite label alphabet $\cY$. An address can be an image--pixel
pair $x=(j,p)$. Evaluation weights $\nu(x)\geq0$ satisfy $\sum_x\nu(x)=1$.
Writing $X\sim\nu$ represents evaluation averaging; it does not imply that
the coordinates themselves move or that views are sampled from $\nu$.

A teacher interpretation $f$ specifies the label distribution
\begin{equation}
 P_f:\cX\longrightarrow\mathcal P(\cY),\qquad
 x\longmapsto P_f(\cdot\mid x).
 \label{eq:target}
\end{equation}
This is what the teacher wants the learner to approximate. A deterministic
segmentation is the special case of a point mass at every address. More
generally, $P_f$ may retain ambiguity even when $f$ is fixed.

To discuss information, consider an ensemble of sessions with $F\sim\mu$.
Since the image collection is fixed, $\mu$ is already conditional on that
collection; random image collections instead require a conditional source
distribution $\mu(\cdot\mid J)$. The random distribution $P_F$ is the
image of $F$ under $f\mapsto P_f$, not another independently chosen object.
For a hypothetical evaluation draw, whether $F$ is discrete or continuous,
\begin{equation}
 P(X=x,Y=y\mid F=f)=\nu(x)P_f(y\mid x).
 \label{eq:generative}
\end{equation}
Here $X$ is chosen independently for evaluation. Actual correction addresses
are chosen by the interaction and generally do not have distribution $\nu$.

The source distribution $\mu$, the target $P_f$, the learner's predictive
distribution $Q_i$, and its beliefs over candidate models are different
objects. In particular, the learner need not know $\mu$ or $P_f$.

\paragraph{Scope of the mathematical comparison}
The formal core concerns pointwise prediction on a fixed collection; it is
not a common probabilistic model for every application reviewed.
Table~\ref{tab:scope} makes the required extensions explicit. Random view
selection within the collection is already represented and does not require
an unseen-image model.

\begin{table}[htbp]
\centering
\caption{Scope of the formal core and limits of transfer.}
\label{tab:scope}
\begin{tabularx}{\linewidth}{@{}>{\raggedright\arraybackslash}p{0.29\linewidth}>{\raggedright\arraybackslash}X@{}}
\toprule
Setting or claim & Representation or additional requirement\\
\midrule
Fixed-pool additive log loss & Directly represented by $P_f$, $\nu$, and the learner's predictions.\\[3pt]
Joint-mask uncertainty; topology or expected Dice/IoU & Pixel marginals do not determine the joint mask distribution; use a joint target and a suitable loss.\\[3pt]
Generalization to unseen images & Specify an image population and an independent evaluation protocol; random views of the fixed pool are insufficient.\\[3pt]
Continuous target families & Mutual information can be defined through joint/product KL; finite discrete entropy differences require additional restrictions.\\[3pt]
Changing human interpretation & Introduce target dynamics and the evaluation reference; asynchronous training alone does not make the target change.\\
\bottomrule
\end{tabularx}
\end{table}

\subsection{Pre-response and post-response memory}
Let $S_{i-1}$ be the learner state after the first $i-1$ responses, and let
\begin{equation}
 Q_i(\cdot\mid x)=Q(\cdot\mid x,S_{i-1}).
 \label{eq:predictor}
\end{equation}
Thus $Q_i$ generates the round-$i$ display; $Q_{i+1}$ is the readout after
response $i$. This convention avoids confusing a round index with an
optimiser iteration. A general state is not necessarily a set of models;
it may include weights, optimiser variables, pending updates, and a displayed
checkpoint. In the finite example of Figure~\ref{fig:example-setup}, it is a candidate set.

Use $M_i$ for random post-response memory and $\cM_i$ for its realized value.
Before the response, the system selects a view $\mathcal X_i$ and displays
$\mathcal V_i$. Its selector $\Lambda$ may depend on $Q_i$, past memory, and
state; random selection uses protocol randomness. The visualization function
is deterministic conditional on its inputs, e.g.,
$\mathcal V_i=\vis(Q_i,\mathcal X_i)$. The view record identifies the actual
checkpoint and any display settings relevant to the teacher. Define nested
records
\begin{equation}
 \cM_0=\emptyset,\qquad
 \cZ_i=(\cM_{i-1},\mathcal X_i,\mathcal V_i),\qquad
 \cM_i=(\cZ_i,r_i).
 \label{eq:memory}
\end{equation}
Their random counterparts satisfy $M_i=(Z_i,R_i)$. Nesting preserves order,
repetitions, and the distinction between what was displayed and what was
corrected. It is not an unordered union of annotated pixels.

Let $R_i=(\Xi_i,H_i)$ denote a random spatial response, with realization
\begin{equation}
 r_i=(\xi_i,h_i),\qquad \xi_i\subseteq\mathcal X_i,\qquad
 h_i:\xi_i\to\cY.
 \label{eq:response}
\end{equation}
An empty response is permitted and must be recorded if it is to be analysed.
In one concrete logging convention, an explicit submit commits a batch of
label assignments under an unchanged view/checkpoint; ``skip'' commits the
empty batch. Reassigning an address already in the current batch first commits
that batch and starts another. Record timestamps separately; erasures and
navigation need their own action symbols, not background labels. The baseline
in Figure~\ref{fig:response-code} uses only this paint/skip alphabet.
The teacher-response kernel is
\begin{equation}
 R_i\sim\rho_f(\cdot\mid Z_i=\cZ_i).
 \label{eq:teacher}
\end{equation}
Dependence on round number can be recovered from memory. Fixed teacher
characteristics can be included in the kernel specification; unobserved
behavioural variation can be marginalized. Unlike $P_f$, this distribution
describes \emph{where and how the teacher responds to a display}, not just a
label at a fixed address. Its labels need not be independent draws from
$P_f$: strategic selection and annotation noise must be specified separately.
One contradictory label does not eliminate a distributional target whose
response model assigns it positive probability.

The learner then updates its state using $\Phi$, possibly with independent
training randomness. In an asynchronous implementation, $Q_{i+1}$ must refer
to a declared next checkpoint or a declared terminal readout, not necessarily
an immediately converged fit of every received correction. This affects
assimilation and delay, without by itself changing the target $f$.

\paragraph{Mixed initiative and what must be logged}
The system-initiative model chooses the view before the teacher's response.
If the teacher navigates or overrides the view, that action may itself convey
target information. It must either be logged and accounted for as an earlier
teacher message in the causal sequence, or remain as additional information
in $Z_i$. Section~\ref{sec:coding} keeps that term explicit. A response contains
its \emph{marked addresses}, not necessarily the whole displayed view.

Figure~\ref{fig:example-setup} instantiates the interaction on two images
with two pixels each, making the first correction and learner update explicit.

\begin{figure}[!htbp]
\begin{runningexample}{Setup and first correction}
Take two images with two addresses each, $\cY=\{0,1\}$, and
$\cX=\{(a,1),(a,2),(b,1),(b,2)\}$. The shared intensities and target labels are
\begin{center}
\begin{tabular}{lcccc}
\toprule
Address & $(a,1)$ & $(a,2)$ & $(b,1)$ & $(b,2)$\\
\midrule
$J(x)$ & 0.2 & 0.7 & 0.4 & 0.9\\
$P_f(1\mid x)$ & 0 & 0 & 0 & 1\\
\bottomrule
\end{tabular}
\end{center}
Let $P_0,\ldots,P_{15}$ denote all deterministic binary label vectors, numbered
in the displayed address order. Thus $P_f=P_1$. For information calculations,
assume the source is uniform over these 16 targets. Set $S_0=\cC_0=
\{P_0,\ldots,P_{15}\}$, and update by removing candidates inconsistent with
faithful noiseless labels. Address choices in this example do not encode extra
target information: they are independent of the unknown labels conditional on
past history and the view. This assumption is what makes uniform beliefs over
surviving candidates legitimate.

The system first displays image $a$. A predictor can give both labels
probability $1/2$, displayed as foreground under a specified threshold tie rule.
The teacher marks $r_1=\{(a,1)\mapsto0\}$. The record becomes
$\cM_1=((\emptyset,\mathcal X_1,\mathcal V_1),r_1)$ and
$\cC_1=\{P_0,\ldots,P_7\}$. One possible learner is the average
\begin{equation}
 \widetilde Q_{i+1}(y\mid x)
 =\frac{1}{|\cC_i|}\sum_{P\in\cC_i}P(y\mid x).
 \label{eq:uniform-learner}
\end{equation}
It changes the foreground probabilities from $(1/2,1/2,1/2,1/2)$ to
$(0,1/2,1/2,1/2)$. The candidate set is an explicit state only in this small
example; the framework does not require enumerating all segmentations.
\end{runningexample}
\caption{A four-pixel interaction with a uniform source over 16 deterministic
targets and one faithful correction.}
\label{fig:example-setup}
\end{figure}

\section{Evaluation-equivalence and predictive distortion}
\label{sec:distortion}
Define
\begin{equation}
 T_f=\{P_g:P_g(\cdot\mid x)=P_f(\cdot\mid x)
                     \text{ for }\nu\text{-almost every }x\}.
 \label{eq:equivalence}
\end{equation}
$T_f$ is a single equivalence class; $T_F$ is a random class induced by $F$.
Neither is the learner's current candidate set. Changing $\nu$ can change the
target class; changing the learner family does not. Different parameters
yielding the same conditional distributions do not represent different
evaluation targets.

Log loss is a strictly proper predictive loss: in expectation it is minimized
by the data-generating label distribution, where the relevant scores are
well-defined \cite{gneiting2007proper}. For log loss, the fixed-target
cross-entropy risk decomposes as
\begin{align}
 \CE(P_f,Q)
 &= -\sum_x\nu(x)\sum_yP_f(y\mid x)\log_2 Q(y\mid x)\nonumber\\
 &=\underbrace{-\sum_x\nu(x)\sum_yP_f(y\mid x)\log_2P_f(y\mid x)}_{H_f(Y\mid X)}
  +\underbrace{\sum_x\nu(x)\KL(P_f(\cdot\mid x)\Vert Q(\cdot\mid x))}_{\tau(T_f,Q)}.
 \label{eq:distortion}
\end{align}
The algebra is given in \ref{app:risk-derivation}.
The first term is the label-variability floor for that fixed target. The second
is predictive distortion, measured in bits per evaluation draw. We use
$0\log 0=0$ and assign infinite loss when $P_f(y\mid x)>0$ but $Q(y\mid x)=0$
at a positively weighted address. Finite-risk calculations assume positivity
on the target's support. Distortion is well-defined on $T_f$ because its
members agree wherever $\nu$ assigns weight.

For a learner family $\mathcal Q$, the approximation floor is
\begin{equation}
 \tau_{\mathcal Q}^{\inf}(T_f)=\inf_{Q\in\mathcal Q}\tau(T_f,Q).
 \label{eq:approximation}
\end{equation}
It is distinct from $H_f(Y\mid X)$ and from optimisation or finite-data error.
If $P_f\notin\mathcal Q$, the infimum need not be attained and can be positive
(or infinite). More labels do not automatically make an arbitrary update
reduce distortion.

Figure~\ref{fig:example-distortion} applies these definitions to the first
correction in Figure~\ref{fig:example-setup}. Changing only the evaluation
weights changes both the equivalence class and the measured improvement.

\begin{figure}[!htbp]
\begin{runningexample}{Evaluation-equivalence and distortion}
In the four-pixel example, the label-variability floor is zero. With uniform
$\nu$, $T_f=\{P_1\}$ and distortion falls from $1$ to $3/4$ after $r_1$. If
each address in image $a$ has weight $1/2$ and image $b$ has weight zero,
$T_f=\{P_0,P_1,P_2,P_3\}$ and distortion instead falls from $1$ to $1/2$.
The larger equivalence class is a less demanding evaluation target, not a
larger remaining candidate set. These values sum over addresses and labels,
not over candidate models. A deterministic target contributes one nonzero
label term at each address, including background labels.
\end{runningexample}
\caption{Evaluation weights determine which targets are equivalent and how
much the same correction reduces distortion.}
\label{fig:example-distortion}
\end{figure}

\section{Information gain, expected improvement, and their differences}
\label{sec:gain}
\subsection{What is uncertain, and under which distribution?}
For the entropy-difference arguments in this review, assume $T_F$ is discrete
with $H(T_F)<\infty$. Initially,
\begin{equation}
 p_{T_F,0}(t)=P(T_F=t),\qquad
 H(T_F)=-\sum_t p_{T_F,0}(t)\log_2p_{T_F,0}(t).
 \label{eq:target-entropy}
\end{equation}
For discrete $F$, $p_{T_F,0}(t)=\sum_{f:T_f=t}\mu(f)$; in general it is
$P(F\in\{f:T_f=t\})$. The entropy sum counts distinct classes, not
interpretations. Only $T_F$, not necessarily $F$, is assumed discrete.

After a realized memory,
\begin{equation}
 p_{T_F,i}(t)=P(T_F=t\mid M_i=\cM_i),\qquad
 H(p_{T_F,i})=H(T_F\mid M_i=\cM_i).
 \label{eq:posterior}
\end{equation}
These probabilities are induced by the source and interaction model. The prior
$\mu$ does not change with the round; conditioning on the history changes the
posterior. They are not automatically the learner's internal beliefs $w$.
The notation is a theoretical description even when computing these
probabilities is impractical.

For a possible response $r$ in the support of the response distribution,
let $\cM_i^r=(\cZ_i,r)$ and define the full-round gain
\begin{equation}
 \Gain_i(r;\cZ_i)=
 H(T_F\mid M_{i-1}=\cM_{i-1})
 -H(T_F\mid M_i=\cM_i^r).
 \label{eq:gain}
\end{equation}
We suppress the context argument when fixed. A realized gain can be negative:
an unexpected response can broaden the posterior. Entropy reduction is not a
distance from the true target or a guarantee of correct belief. The expression
also includes any information supplied by the view and display before the
response. The response-only counterpart replaces its first entropy by
$H(T_F\mid Z_i=\cZ_i)$.

\paragraph{Beyond discrete targets}
Unrestricted distribution-valued targets need not have a finite Shannon
entropy. One may specify a finite-resolution target representation for the
entropy calculations, or define mutual information directly through KL
divergence between joint and product measures, with finiteness conditions.
Replacing the displayed entropies by differential entropies is not a harmless
notation change: the latter depend on parameterization and do not provide the
same discrete uncertainty interpretation. We make no finite-gain claim for
arbitrary continuous targets here.

\subsection{Fixed-target expectations versus source averages}
Fix $f$, a pre-response record $\cZ_i$, and a current predictor $Q_i$. For a
deterministic update, let $Q_{i+1}(r)$ be the next predictor. Then
\begin{align}
 \E[\Gain_i(R_i)\mid F=f,Z_i=\cZ_i]
 &=\sum_r\rho_f(r\mid\cZ_i)\Gain_i(r;\cZ_i),
 \label{eq:fixed-gain}\\
 \E[\tau(T_f,Q_{i+1})\mid F=f,Z_i=\cZ_i]
 &=\sum_r\rho_f(r\mid\cZ_i)\tau(T_f,Q_{i+1}(r)).
 \label{eq:fixed-distortion}
\end{align}
For random updates the second summand includes the conditional expectation
over update randomness. If $Z_i$ does not fix the current predictor, also
condition on its realized state. Averaging before a random view is selected
requires an outer expectation over the context distribution. This context
distribution is part of the acquisition protocol, not the evaluation weights
$\nu$.

The fixed-target expected gain in \eqref{eq:fixed-gain} is not in general a
mutual information, and need not be nonnegative. Mutual information appears
when targets and responses are averaged under the same joint source model,
as derived in Section~\ref{sec:coding}. This distinction prevents treating a
teacher's known-$f$ response distribution as the learner's predictive mixture.

For a fixed target and finite expected log losses, the entropy floor cancels:
\begin{equation}
 \CE(P_f,Q_i)-\E[\CE(P_f,Q_{i+1})]
 =\tau(T_f,Q_i)-\E[\tau(T_f,Q_{i+1})],
 \label{eq:risk-reduction}
\end{equation}
where both expectations use the same fixed-target context. There is no
general equality between either side and target information gain.

\subsection{When information improves prediction}
Fix $Z_i=z$; if this record does not determine the current learner state,
condition on that state throughout as well. Draw $X\sim\nu$ independently
of $(F,R_i)$, then a fresh $Y\sim P_F(\cdot\mid X)$ independently of $R_i$
conditional on $(F,X)$. Assume the expected log losses below are finite.
Under the correct joint source model, the Bayes predictors are
\begin{align}
 Q_i^{\mathrm B}(y\mid x)&=P(Y=y\mid X=x,Z_i=z),\nonumber\\
 Q_{i+1}^{\mathrm B}(y\mid x,R_i)&=P(Y=y\mid X=x,Z_i=z,R_i).
 \label{eq:bayes-predictors}
\end{align}
The expected improvement from the Bayes update is
\begin{equation}
 \E[\tau(T_F,Q_i^{\mathrm B})-\tau(T_F,Q_{i+1}^{\mathrm B})\mid Z_i=z]
 =I(Y;R_i\mid X,Z_i=z)\geq0.
 \label{eq:predictive-information}
\end{equation}
For an actual learner based on the same records, define its nonnegative
excess log-loss risks relative to these Bayes predictors:
\begin{align*}
 a_i&=\E\!\left[\KL\bigl(Q_i^{\mathrm B}(\cdot\mid X)
                 \Vert Q_i(\cdot\mid X)\bigr)\mid Z_i=z\right],\\
 a_{i+1}&=\E\!\left[\KL\bigl(Q_{i+1}^{\mathrm B}(\cdot\mid X,R_i)
                 \Vert Q_{i+1}(\cdot\mid X)\bigr)\mid Z_i=z\right].
\end{align*}
The expectations also average fresh independent update randomness, if used.
Adding these excess risks to the Bayes losses gives
\begin{equation}
 \E[\tau(T_F,Q_i)-\tau(T_F,Q_{i+1})\mid Z_i=z]
 =I(Y;R_i\mid X,Z_i=z)+a_i-a_{i+1}.
 \label{eq:assimilation-gap}
\end{equation}
Prediction therefore improves on average precisely when the predictive
information exceeds the increase $a_{i+1}-a_i$ in excess risk. An update can
forgo this benefit by fitting corrections poorly or damaging predictions
elsewhere; reducing existing excess risk can instead add to the information
benefit. \ref{app:predictive-information} gives the derivation.
Both identities average over targets and responses under the same conditional
source model; they do not guarantee improvement for each target or response.
The information term is response-only predictive information, bounded above
by $I(T_F;R_i\mid Z_i=z)$; neither is the full-round realized gain.

BIFSeg's comparison from identical initial predictions and scribbles shows
that different uses of the same corrections can yield different current-mask
Dice scores \cite{wang2018bifseg}. The fixed-stream replay proposed in
Section~\ref{sec:research} similarly isolates assimilation from changed
responses. These outcome comparisons do not estimate the information term
in \eqref{eq:assimilation-gap}.

For other losses, an optimal decision can ignore the response, so its expected
risk cannot increase if the available decisions are unchanged. The equality
with mutual information is specific to log loss. Dice or topology requires
the corresponding joint-mask decision model, not merely pixel marginals.

Figure~\ref{fig:example-response} evaluates the next round under a
history-insensitive response policy. It uses the source and learner of
Figure~\ref{fig:example-setup} with uniform evaluation weights, and averages
over the next view and correction.

\begin{figure}[!htb]
\begin{runningexample}{A history-insensitive teacher}
Continue from $r_1=\{(a,1)\mapsto0\}$ with the uniform candidate-average
predictor, uniform evaluation weights, and the fixed target $P_1$. The system
selects image $a$ or $b$ with probability $1/2$. In the selected image, a
history-insensitive teacher selects its first pixel with probability
$1-\alpha$, or selects either pixel uniformly with probability $\alpha$:
\begin{equation}
 \rho_f^{(\alpha)}=(1-\alpha)\rho_f^{\mathrm{first}}
                  +\alpha\rho_f^{\mathrm{uniform}},\qquad 0\leq\alpha\leq1.
 \label{eq:alpha}
\end{equation}
Labels are faithful and repeatable. The following probabilities marginalize
over the random view; they are not conditional on one selected image.
\begin{center}
\begin{tabular}{lccc}
\toprule
Response $r$ & $P(R_2=r\mid F=f,M_1=\cM_1)$ & $\Gain_2$ & $\tau(T_f,\widetilde Q_3)$\\
\midrule
$(a,1)\mapsto0$ & $1/2-\alpha/4$ & 0 & $3/4$\\
$(a,2)\mapsto0$ & $\alpha/4$ & 1 & $1/2$\\
$(b,1)\mapsto0$ & $1/2-\alpha/4$ & 1 & $1/2$\\
$(b,2)\mapsto1$ & $\alpha/4$ & 1 & $1/2$\\
\bottomrule
\end{tabular}
\end{center}
Thus, conditional on the first record and fixed $f$ but averaging over the
next view and response,
\begin{equation}
 \E[\Gain_2]=\frac12+\frac\alpha4,\qquad
 \E[\tau(T_f,\widetilde Q_3)]=\frac58-\frac\alpha{16}.
 \label{eq:toy-expectations}
\end{equation}
Here every newly labelled pixel contributes one target bit and removes $1/4$
bit of distortion. This proportionality follows from the deliberately
independent, uniform, noiseless example, not from a general theorem.
$\alpha$ changes the response distribution and hence the distribution of
future predictors; it does not change $\nu$ or $P_f$.
\end{runningexample}
\caption{Expected gain and distortion after one more correction depend on
the teacher's response policy, even with faithful labels.}
\label{fig:example-response}
\end{figure}

\FloatBarrier
The proportionality in Figure~\ref{fig:example-response} is special to its
assumptions. Figure~\ref{fig:example-relevance} changes evaluation relevance
while keeping the learner calibrated; Figure~\ref{fig:example-updates}
instead changes the learner's predictions and specifies its update.
Both comparisons separate target information from predictive benefit.

\begin{figure}[!htbp]
\begin{runningexample}{Equal information, unequal evaluation relevance}
Keep the uniform source and candidate-average learner after $r_1$, but set
$\nu(a,1)=1/4$, $\nu(a,2)=1/2$, and $\nu(b,1)=\nu(b,2)=1/8$.
Every weight is positive, so the evaluation-equivalence classes are unchanged.
The predictor is calibrated to the source posterior and its distortion is
$3/4$. For an unknown independent fair label, observing it gives one target
bit while removing $\nu(x)$ bits of evaluation-weighted distortion:
\begin{center}
\begin{tabular}{lccc}
\toprule
Correction & Gain & Post-update distortion & Distortion reduction\\
\midrule
$(a,2)\mapsto0$ & 1 & $1/4$ & $1/2$\\
$(b,2)\mapsto1$ & 1 & $5/8$ & $1/8$\\
\bottomrule
\end{tabular}
\end{center}
Both responses cost one mark. Different predictive benefits here arise from
evaluation relevance, not from miscalibrated beliefs or unsuccessful fitting.
Figure~\ref{fig:example-updates} restores uniform $\nu$ and varies the learner instead.
\end{runningexample}
\caption{Equal information gain can give different distortion reductions
when evaluation weights differ.}
\label{fig:example-relevance}
\end{figure}

\begin{figure}[!htb]
\begin{runningexample}{Alternative learner updates}
Keep the uniform source posterior on $\cC_1$ but choose a different learner
predictor. To make the update fully specified, let the learner maintain
independent Bernoulli beliefs at the three unknown pixels. A correction sets
its corresponding probability to the supplied hard label and leaves all other
probabilities unchanged. These learner beliefs differ from the uniform source
posterior used to calculate gain. This is a local illustrative variant, not
the uniform update of \eqref{eq:uniform-learner}.

First take $Q_2(1\mid x)=(0,1/2,1/2,1/4)$, giving initial distortion $1$.
\begin{center}
\begin{tabular}{lccc}
\toprule
Correction & Gain & Post-update distortion & Distortion reduction\\
\midrule
$(a,2)\mapsto0$ & 1 & $3/4$ & $1/4$\\
$(b,2)\mapsto1$ & 1 & $1/2$ & $1/2$\\
\bottomrule
\end{tabular}
\end{center}
The two labels reveal equal target information but repair predictions of
different quality. Now take $Q_2(1\mid x)=(0,7/8,1/2,1/2)$, giving initial
distortion $5/4$. Compare two-mark responses:
\begin{center}
\begin{tabular}{lccc}
\toprule
Correction & Gain & Post-update distortion & Distortion reduction\\
\midrule
$(b,1)\mapsto0,(b,2)\mapsto1$ & 2 & $3/4$ & $1/2$\\
$(a,1)\mapsto0,(a,2)\mapsto0$ & 1 & $1/2$ & $3/4$\\
\bottomrule
\end{tabular}
\end{center}
The second response repeats one known label but fixes the more serious
remaining prediction error. Both have equal marked-location work. For example,
the probability of the true background label at $(a,2)$ was $1/8$, contributing
$-\tfrac14\log_2(1/8)=3/4$ to initial distortion. Address choices are fixed
comparisons, not a separate channel encoding $f$. These examples demonstrate
the distinction between acquisition and assimilation; they do not demonstrate
superiority of any real acquisition method.
\end{runningexample}
\caption{More target information need not produce a larger distortion
reduction when learner beliefs differ from the source posterior.}
\label{fig:example-updates}
\end{figure}

% Keep the example equations and figures before the next theoretical section.
\FloatBarrier
\subsection{When does this perspective recover BALD?}
\label{sec:bald-link}
The standard expected-KL form of \eqref{eq:bald} is
\begin{equation}
 \operatorname{BALD}(x)=\sum_Pw(P\mid\cM)
 \KL\bigl(P(\cdot\mid x)\Vert\widehat Q(\cdot\mid x,\cM)\bigr).
 \label{eq:bald-kl}
\end{equation}
\ref{app:bald-derivation} gives the expansion.
BALD averages model disagreement at a proposed address. Distortion instead
averages discrepancy from the fixed teacher target over evaluation addresses.
The two KL expressions have different averaging distributions and meanings.
Equation~\eqref{eq:predictive-information} identifies the distinction:
BALD values information about the candidate model; predictive improvement
values information about labels at evaluation addresses.

For exact agreement with response information about $T_F$, the learner's
beliefs must match the source-conditioned beliefs, the response must be the
requested label under the same observation model, and label probabilities
must depend on the candidate only through its evaluation class. To see the
last condition, let $\mathsf P$ be the random candidate model and
$T=T(\mathsf P)$ its class. Then
\begin{equation}
 I(\mathsf P;Y\mid x,\cM)
 =I(T;Y\mid x,\cM)+I(\mathsf P;Y\mid T,x,\cM).
 \label{eq:quotient-bald}
\end{equation}
The last term vanishes at an evaluation-relevant address when members of each
class specify the same label probabilities. It need not vanish for queries
outside the support of $\nu$. Unknown candidate parameters are not always
evaluation-relevant unknowns.

Figure~\ref{fig:example-bald} specializes the comparison to the four-pixel
setting. Its difference in expected gain also reflects different available
actions, not only a different acquisition criterion.

\begin{figure}[!htbp]
\begin{runningexample}{Comparison with BALD}
In the uniform four-pixel example all surviving candidates are deterministic,
so BALD is zero at $(a,1)$ and one bit at every other pixel. A requested new
pixel gives gain one and post-update distortion $1/2$. The history-insensitive
teacher of \eqref{eq:alpha} gives \eqref{eq:toy-expectations} instead. This
comparison changes the action space: unrestricted pixel requests are compared
with random views and a teacher who may repeat a mark. It is not evidence that
BALD is intrinsically better than teacher modelling. A fair experiment must
match views, response opportunities, work, and learner updates.
\end{runningexample}
\caption{The four-pixel BALD comparison changes both the selection rule and
the response opportunities; it is not a matched-policy experiment.}
\label{fig:example-bald}
\end{figure}

\section{Human work and teaching objectives}
\label{sec:work}
For responses with random supports $\Xi_i$, define marked-location work
\begin{equation}
 W_n=\sum_{i=1}^n|\Xi_i|.
 \label{eq:work}
\end{equation}
Repeated marks in different rounds count again, but support sets do not
measure retracing, inspection, navigation, or time. For human effort, specify
$C_n=\sum_{i=1}^n c_i$ in a declared unit: $c_i=|\Xi_i|$ recovers $W_n$;
active or elapsed seconds measure different resources. Record repeats, edits,
empty responses, and waiting without double-counting overlapping activities.
An empty response can cost time. Navigation and inspection must be held fixed
or charged if a policy changes them; otherwise target information can enter
through an uncharged action.

Fix a learner, a horizon $n$, an evaluation distribution, and a class $\Pi$
of allowed interaction policies. A policy specifies which choices are
controlled by the system and which by the teacher. Two ideal objectives are
\begin{align}
 \sup_{\pi\in\Pi}\quad&
 \E_\pi\!\left[\sum_{i=1}^n\Gain_i(R_i;Z_i)\right]
 &&\text{subject to }\E_\pi[C_n]\leq b,
 \label{eq:gain-objective}\\
 \inf_{\pi\in\Pi}\quad&\E_\pi[\tau(T_F,Q_{n+1})]
 &&\text{subject to }\E_\pi[C_n]\leq b.
 \label{eq:distortion-objective}
\end{align}
Expectations average over targets and interaction and training randomness
under the declared protocol. For a single fixed target, condition throughout
on $F=f$ instead. Gain and its posterior are defined under the joint model
induced by each policy, not by applying one policy's posterior to another
policy's responses. Suprema and infima need not be attained. These are
definitions of objectives, not claims that their values or optimizers can be
computed from logs. An expected-cost constraint does not bound the effort in
every individual session; a hard budget needs an explicit termination rule.
The predictor at the horizon must have a declared readout
and computational budget. These are fixed-horizon objectives; adaptive
stopping requires a stopping-time formulation and accounting for a
teacher-controlled stop signal.

Comparisons at a common budget additionally require a common action set,
initialization, evaluation distribution, and declared training resource.
Alternatively, changes in these components must be explicit experimental
factors. A more powerful annotation interface or extra fitting iterations
can explain a benefit attributed to a query score. Neither equal mark counts
nor a ratio labelled ``information per click'' removes those alternatives.

Gain and log-loss distortion are measured in bits, but are not the same
quantity: one concerns target uncertainty, the other error per evaluation
draw. Work is measured in marks, or separately in seconds. Constraints such
as \eqref{eq:gain-objective} avoid adding incompatible units. Local ratios
such as expected distortion reduction divided by expected work are useful
diagnostics, but are not generally optimal multi-round policies. A ratio of
expectations is also different from an expectation of a ratio, particularly
when responses can be empty.

\section{Prequential evaluation: what response predictability establishes}
\label{sec:mdl}
For a discrete alphabet and integer lengths satisfying Kraft's inequality,
the shortest expected prefix-code length lies between entropy and entropy
plus one bit. Ideal lengths $-\log_2 q(a)$ instead correspond to a chosen
probability model; arithmetic coding can implement a sequence code with
small additional overhead. These ideal lengths need not be integers.

Two-part MDL describes a model and then data conditional on that model.
Prequential coding predicts and encodes observations using only already
decoded observations \cite{rissanen1978modeling,grunwald2007mdl,
grunwald2020revisited}. This instantiates the sequential forecast evaluation
advocated by \cite{dawid1984prequential}, with the logarithmic scoring rule
\cite{gneiting2007proper}. For $D_N=(d_1,\ldots,d_N)$,
\begin{align}
 \ell_{\mathrm{two}}(\theta,D_N)&=\ell(\theta)+\ell(D_N\mid\theta),
 \label{eq:two-part}\\
 \ell_{\mathrm{preq}}(D_N)&=-\sum_{t=1}^N
       \log_2 q_t(d_t\mid d_1,\ldots,d_{t-1}).
 \label{eq:prequential}
\end{align}
For continuous parameters, a density at an exact $\theta$ is not a probability
mass and $-\log_2 p(\theta)$ is not by itself a parameter code. A two-part
code must specify a finite precision and a decodable parameter description.

The comparative use here is forecasting a fixed response stream before seeing
each response, not minimizing how much useful supervision a teacher supplies.
For well-specified i.i.d.\ data in a fixed regular $k$-parameter family,
suitable two-part and Bayesian-mixture codes have leading redundancy
$(k/2)\log_2 N$ relative to fitted likelihood. Regularity requires an interior
parameter, nonsingular Fisher information, and concentration; an expected
expansion also requires control of the remainder (\ref{app:mdl}). This is
not an arbitrary neural-training guarantee or a bound on human effort.
Adaptive responses do not justify replacing $N$ by rounds or marks.

\section{Information in adaptive response streams}
\label{sec:coding}
Fix the shared initial image collection, evaluation weights, protocol, and
initial state; all probabilities in this section condition on them. The
initial memory is empty. Let $\widehat\rho_i(r\mid z)$ be a normalized causal
response model, shared by encoder and decoder \emph{before} the response is
encoded. It must not depend on the current unknown response or the hidden
target except through available information. Define
\begin{equation}
 \ell_i=-\log_2\widehat\rho_i(R_i\mid Z_i),\qquad
 \ell_{\mathrm{PT}}(n)=\E\!\left[\sum_{i=1}^n\ell_i\right].
 \label{eq:pt-length}
\end{equation}
This codes a correction process, including support selection, rather than
only the target label at a requested pixel. The coder and decoder must agree
on context and coding probabilities; they need not independently train
identical segmentation networks. If obtaining a context requires additional
communication, its cost is outside \eqref{eq:pt-length} unless included
explicitly. In particular, the response code is not a code for the entire
interactive system's traffic.

The true context-conditional response mass is
$\rho_i(r\mid z)=P(R_i=r\mid Z_i=z)$, obtained by marginalizing the
target-conditioned kernels under the source posterior. The estimated coder
$\widehat\rho_i$ is neither $P_f$ nor $Q_i$ and need not be statistically
consistent. Assume it is positive wherever the true response mass is
positive when a finite length is desired.

\begin{proposition}[Response coding and information]
\label{prop:coding}
For discrete responses, shared decoding context, a normalized causal
$\widehat\rho_i$, and finite terms,
\begin{align}
 \E[\ell_i]
 &=H(R_i\mid Z_i)
   +\E_{Z_i}\KL(\rho_i(\cdot\mid Z_i)\Vert
                         \widehat\rho_i(\cdot\mid Z_i)),
 \label{eq:code-mismatch}\\
 \ell_{\mathrm{PT}}(n)
 &\geq\sum_{i=1}^nH(R_i\mid Z_i)
 \geq\sum_{i=1}^n I(T_F;R_i\mid Z_i).
 \label{eq:code-bound}
\end{align}
\end{proposition}
These are the standard cross-entropy decomposition and entropy bound on
mutual information, applied to context-conditional responses; the algebra
is in \ref{app:code-derivation}. Their relevance is that they
separate response information from coding-model error.

The residual $H(R_i\mid T_F,Z_i)$ can include support-choice variability,
annotation noise, and behaviour that distinguishes interpretations within
one target class. It is not generally the label-variability floor
$H_f(Y\mid X)$. Combining the identities gives a useful diagnostic
decomposition:
\begin{equation}
 \E[\ell_i]=I(T_F;R_i\mid Z_i)+H(R_i\mid T_F,Z_i)
 +\E_{Z_i}\KL(\rho_i\Vert\widehat\rho_i).
 \label{eq:three-code-terms}
\end{equation}
Long code can mean useful information, response variability, or a poor coder.
Short code can mean good prediction or repeated, uninformative corrections.
Neither minimizing nor maximizing code length alone optimizes teaching.

\subsection{A measurable comparison on the same interaction stream}
\label{sec:prequential-comparison}
Compare two normalized causal models, $\widehat\rho_i^{(0)}$ and
$\widehat\rho_i^{(1)}$, on the same recorded contexts and responses.
Their forecasts must be fixed before the scored response arrives, with
fitting and model selection restricted to prior data or separate development
sessions. Condition throughout this comparison on the shared model and
update specifications, development information, and fitting randomness fixed
before scoring begins. Their description costs are excluded: this is a
conditional predictive-code comparison, not a comparison of complete MDL
descriptions. Assume both
response models have finite expected log loss, so their expected contrast
does not subtract infinite quantities.
Use identical response alphabets and batch boundaries. The observed length
advantage of model 1 is
\begin{equation}
 \Delta\ell_n=\sum_{i=1}^n
 \log_2\frac{\widehat\rho_i^{(1)}(r_i\mid\cZ_i)}
                 {\widehat\rho_i^{(0)}(r_i\mid\cZ_i)}.
 \label{eq:score-comparison}
\end{equation}
This score is computable without $\mu$. Its source expectation satisfies
\begin{equation}
 \E[\Delta\ell_n]=\sum_{i=1}^n\E_{Z_i}
 \left[\KL(\rho_i\Vert\widehat\rho_i^{(0)})
       -\KL(\rho_i\Vert\widehat\rho_i^{(1)})\right],
 \label{eq:expected-score-comparison}
\end{equation}
with all response distributions conditional on the same $Z_i$. This follows
by cancelling the common entropy in \eqref{eq:code-mismatch}; its individual
KL terms need not be observable. History, overlay, or support-geometry features
can be compared this way, holding the conditional label model fixed when
testing support prediction. A positive score establishes better prediction
of that stream, not identification of true gain, improved segmentation, or
the value of a different acquisition policy that would change the stream.
Figure~\ref{fig:response-code} supplies a normalized baseline for this test.

\begin{figure}[!htbp]
\begin{runningexample}{A complete response-code baseline}
For a displayed set of $m$ addresses, draw the batch size uniformly from
$0,\ldots,m$, then a uniform subset of that size and independent uniform
labels. For every admissible paint/skip response,
\begin{equation}
 \widehat\rho_i^{(0)}(\xi,h\mid z)
 =\frac{1}{(m+1)\binom{m}{|\xi|}|\cY|^{|\xi|}}.
 \label{eq:response-baseline}
\end{equation}
There are $\binom{m}{k}|\cY|^k$ responses of size $k$, so their total mass
is $1/(m+1)$. This includes the unique empty response and sums to one.

In image $a$ of the four-pixel example, $m=2$ and $|\cY|=2$.
Skipping has probability $1/3$; each specified one- or two-pixel response
has probability $1/12$. Thus $\{(a,1)\mapsto0\}$ costs $\log_2 12$ bits
even if its label is already known. A causal history-aware forecast can
shorten that code without making the repeated label informative.

This intentionally simple reference models neither stroke geometry nor
teacher optimality. Learned size, support, and label factors can replace
its uniform factors; score each forecast before updating it. Independence
is a baseline choice, not an assumption about real corrections.
\end{runningexample}
\caption{A normalized response model makes same-stream scoring operational
without estimating the teacher's target or information gain.}
\label{fig:response-code}
\end{figure}

\subsection{Which information belongs to the response?}
For nested records with empty initial memory, discrete $T_F$ of finite
entropy, and the joint source--interaction distribution,
the full-round gain in \eqref{eq:gain} satisfies
\begin{equation}
 \E\!\left[\sum_{i=1}^n\Gain_i(R_i;Z_i)\right]
 =I(T_F;M_n).
 \label{eq:gain-telescope}
\end{equation}
The conditional-entropy averaging and telescoping steps are given in
\ref{app:gain-derivation}. This expectation is not the fixed-$f$
average of \eqref{eq:fixed-gain}. Nor is the resulting information necessarily
all conveyed by the responses: the displayed context may contain teacher
choices or other target-dependent information.

Because $Z_i$ contains $M_{i-1}$ and $M_i=(Z_i,R_i)$, the chain rule yields
\begin{align}
 H(T_F\mid M_{i-1})-H(T_F\mid M_i)
 &=I(T_F;Z_i\mid M_{i-1})+I(T_F;R_i\mid Z_i),
 \label{eq:context-split}\\
 I(T_F;M_n)
 &=\sum_{i=1}^n I(T_F;Z_i\mid M_{i-1})
   +\sum_{i=1}^n I(T_F;R_i\mid Z_i).
 \label{eq:full-information}
\end{align}
One can verify the first line by adding and subtracting
$H(T_F\mid Z_i)$; the second telescopes.

\begin{proposition}[When response length bounds full-memory gain]
\label{prop:full-gain}
Under the preceding coding and finite-entropy conditions, suppose the
pre-response context supplies no fresh target information:
\begin{equation}
 I(T_F;Z_i\mid M_{i-1})=0\quad\text{for every }i.
 \label{eq:no-fresh-context}
\end{equation}
Then
\begin{equation}
 \ell_{\mathrm{PT}}(n)\geq I(T_F;M_n)
 =\E\!\left[\sum_{i=1}^n\Gain_i(R_i;Z_i)\right].
 \label{eq:gain-code-bound}
\end{equation}
\end{proposition}
This follows by setting the context terms in \eqref{eq:full-information} to
zero and applying \eqref{eq:code-bound}; see also
\ref{app:gain-derivation}.

A sufficient protocol generates views and displays from the logged past,
fixed initial setup, and fresh target-independent randomization, with no
additional hidden target-dependent input. Such views can depend on a learned
$Q_i$; conditioning on the relevant past is what matters. Teacher navigation
can violate \eqref{eq:no-fresh-context}. Either code that choice in causal
order as a teacher message, or keep its context-information term. Neither
proposition identifies $M_n$ with $\mathbf R_n=(R_1,\ldots,R_n)$, and the
argument does not require reconstructing a random view from response
addresses alone. This is the role of the nested pre-/post-response memory.

\subsection{Support, labels, and work}
The response-code chain rule is
\begin{equation}
 \widehat\rho_i(\xi,h\mid z)
 =\widehat\rho_i(\xi\mid z)\widehat\rho_i(h\mid\xi,z).
 \label{eq:response-factor}
\end{equation}
It assumes no independence. Consequently,
\begin{equation}
 \ell_{\mathrm{PT}}(n)=
 \E\!\left[-\sum_{i=1}^n\log_2\widehat\rho_i(\Xi_i\mid Z_i)\right]
 +\E\!\left[-\sum_{i=1}^n\log_2\widehat\rho_i(H_i\mid\Xi_i,Z_i)\right].
 \label{eq:support-label-code}
\end{equation}
In a shared ordering of marked addresses, the conditional label code factors
into $W_n$ sequential label terms (\ref{app:support-derivation}). Labels may be dependent; no retraining
after each within-response label is required. Support size and addresses
have their own cost. Thus work counts marked locations while coding weights
events by predictability. Pixel counts are not independent sample counts.

There is a parallel information decomposition,
\begin{equation}
 I(T_F;R_i\mid Z_i)
 =I(T_F;\Xi_i\mid Z_i)+I(T_F;H_i\mid\Xi_i,Z_i).
 \label{eq:support-information}
\end{equation}
This support-first decomposition is order-dependent, not a causal attribution:
redundant information enters the first term. Selected addresses can be
informative without their information being estimable or useful to the learner.
The unmarked complement is determined by the view and support, not an
independent channel to count again.

Figure~\ref{fig:example-support} isolates support information using a known
selection convention. It changes the four-pixel source and teacher policy
explicitly, showing a possibility rather than assuming typical human behaviour.

\begin{figure}[!htbp]
\begin{runningexample}{The address can distinguish targets when labels do not}
Use the four addresses with positive evaluation weights, but restrict the
source to equal probabilities on $P_0=(0,0,0,0)$ and $P_1=(0,0,0,1)$.
Display image $a$ under the same initial predictor. Specify a known teacher
policy: under $P_0$, mark $(a,1)\mapsto0$; under $P_1$, mark $(a,2)\mapsto0$.
Both targets agree on both displayed labels. Nevertheless,
\[
 H(T_F\mid Z)=1,\qquad I(T_F;\Xi\mid Z)=1,\qquad
 I(T_F;H\mid\Xi,Z)=0.
\]
The address identifies the target; its label adds no information once the
address is known. A learner using only compatibility with the supplied label
retains both candidates. This is a deliberately explicit signalling convention,
not a model of typical human behaviour: without a known or independently
validated selection mechanism, the same inference is unjustified. It isolates
the possibility expressed by \eqref{eq:support-information} from the empirical
question of whether real support choices can be exploited.
\end{runningexample}
\caption{Under a known selection convention, the chosen address identifies
the target even though both targets agree on the supplied labels.}
\label{fig:example-support}
\end{figure}

\paragraph{Information acquired is not necessarily retained}
If the terminal predictor is a function of $M_n$ (or a stochastic readout
conditionally independent of $T_F$ given $M_n$), data processing gives
$I(T_F;Q_{n+1})\leq I(T_F;M_n)$. This is a statement about information,
not an upper or lower bound on predictive distortion without further
assumptions. If the learner has additional target-dependent side information,
it must be included in the conditioning or the state. The same attention
to reconstruction and side information is needed in rate--distortion
formulations \cite{nokleby2015ratedistortion}. Improving with every prefix
does not establish optimal successive refinement \cite{equitz1991successive}.

\section{What does an improvement per human effort establish?}
\label{sec:research}
\subsection{A map of objectives rather than a single efficiency score}
Table~\ref{tab:objectives} summarizes the distinctions developed above.
Methods are related by their objective structure, not asserted to optimize
identical quantities. The cited applications extend beyond segmentation;
the specialization here is the corrective spatial interaction.

\begin{table}[htbp]
\centering
\caption{Quantities and their relation to established approaches.}
\label{tab:objectives}
\begin{tabularx}{\linewidth}{@{}>{\raggedright\arraybackslash}p{0.22\linewidth}>{\raggedright\arraybackslash}X>{\raggedright\arraybackslash}X@{}}
\toprule
Quantity & Interpretation & Related approach and qualification\\
\midrule
Expected gain & Target information acquired & BALD/BatchBALD under matching beliefs, observation models, and target relevance \cite{houlsby2011bald,kirsch2019batchbald}.\\[3pt]
Expected distortion & Predictive error & Variance reduction for squared error under negligible bias, not general KL \cite{cohn1996active}.\\[3pt]
Gain/work & Information per declared cost unit & Cost-aware acquisition; CEREALS uses informativeness and predicted annotation cost, not exact target gain \cite{mackowiak2018cereals}.\\[3pt]
Distortion reduction/work & Predictive improvement per declared cost unit & Related to ROI acquisition; Haertel et al. consider accuracy benefit and annotation cost \cite{haertel2015roi}.\\[3pt]
Prequential length & Predictability of a response stream & Applies sequential predictive evaluation to support and labels \cite{dawid1984prequential,blier2018description}.\\[3pt]
Length/work & Code bits per declared cost unit & A response-model diagnostic, not a measure of target gain.\\
\bottomrule
\end{tabularx}
\end{table}

For comparing coding models, hold the response stream fixed and predict each
response before fitting it. A short code on a different stream does not
establish a better teaching policy. Conversely, comparing policies requires
evaluating the resulting predictions and work, not just their code lengths.
These comparisons separate response predictability, information acquisition,
and learning benefit.

\subsection{Observed, estimated, and latent quantities}
Table~\ref{tab:observability} separates what the mathematical definitions
permit from what an interaction log identifies. Fitting a statistical response
predictor does not necessarily identify the human's attention or selection
mechanism. Likewise, exact target gain need not be an empirical endpoint in
order to test improved learning.

\begin{table}[htbp]
\centering
\caption{Evidence required to operationalize the framework.}
\label{tab:observability}
\begin{tabularx}{\linewidth}{@{}>{\raggedright\arraybackslash}p{0.25\linewidth}>{\raggedright\arraybackslash}X>{\raggedright\arraybackslash}X@{}}
\toprule
Quantity & Available evidence & What does not follow\\
\midrule
View, checkpoint, support, labels, times & Direct logs, if the interface records them. & Display is not inspection; non-correction is not acceptance.\\[3pt]
$W_n$ and measured $C_n$ & Support counts or a declared action/timing protocol. & Equal marks need not mean equal human effort.\\[3pt]
Response log loss & Score causal forecasts on the same recorded stream. & A better response predictor need not identify gain or improve learning.\\[3pt]
Predictive risk difference & Representative, independently assessed labels for declared checkpoints. & Training corrections alone need not represent the evaluation distribution.\\[3pt]
Source posterior, exact target gain, entropy floor & A specified source and observation model or additional identifying evidence. & A single correction log generally does not identify these quantities.\\[3pt]
Responses to undisplayed views & Interventions, or justified overlap and response-model assumptions. & Replaying recorded corrections does not reveal counterfactual human choices.\\
\bottomrule
\end{tabularx}
\end{table}

For fixed checkpoints $Q^{(0)}$ and $Q^{(1)}$, representative audit labels
give a particularly useful bridge. With independent draws
$X_j\sim\nu$, $Y_j\sim P_f(\cdot\mid X_j)$, independent of fitting and
checkpoint selection, define
\begin{align}
 \widehat\Delta_m&=\frac1m\sum_{j=1}^m
    \log_2\frac{Q^{(1)}(Y_j\mid X_j)}{Q^{(0)}(Y_j\mid X_j)},
 \label{eq:audit-improvement}\\
 \E[\widehat\Delta_m]&=\tau(T_f,Q^{(0)})-\tau(T_f,Q^{(1)}).
 \label{eq:audit-improvement-meaning}
\end{align}
Assume the relevant log losses are finite. The unknown entropy floor cancels,
so estimating predictive improvement does not require recovering $\mu$ or
absolute distortion. The audit's reference-label mechanism must represent
the declared target; annotator noise or consensus labels can define a
different estimand. Spatial, image, and participant dependence must be
reflected in uncertainty estimates, and choosing a checkpoint on the same
audit requires separate validation or an appropriate sequential procedure.

\subsection{Evidence for learning and efficiency}
A benefit claim must specify the outcome, reference population, cost unit,
allowed actions, and training resources. Independent paired risk comparisons
apply beyond log loss; the KL and mutual-information identities do not.
Dice and topology require evaluating whole masks rather than only marginal
label distributions. Table~\ref{tab:interaction-evidence} distinguishes
current-mask completion, later-image prediction, and response prediction,
as well as human time from cost proxies.

Three questions organize the research priorities below: what selection
reveals, whether using it improves assimilation, and whether the resulting
interaction policy improves outcomes at matched effort. They require
different evidence. A useful response predictor need not identify human
intentions, but interpreting non-correction as acceptance requires more than
predictive accuracy.

\subsection{What does selection reveal?}
Pedagogical and teacher-aware learning interpret why an example was chosen
\cite{shafto2014rational,yuan2021teacheraware}; RClicks models click locations
\cite{antonov2024rclicks}. For spatial corrections, first test whether history
and display improve response forecasts on a common recorded stream
\eqref{eq:score-comparison}. This tests prediction of behaviour, not the
correctness of unmarked locations.

In particular, displayed, inspected, and accepted are different events.
An unmarked region can be correct, overlooked, ambiguous, or deferred. A
simple identification problem illustrates the limitation. At a randomly
audited displayed location, let $A$ indicate an actual prediction error and
$C$ indicate a corrective mark. Under the restrictive assumption that only
errors are marked,
\begin{equation}
 P(C=1\mid z)
 =P(A=1\mid z)P(C=1\mid A=1,z).
 \label{eq:selection-identification}
\end{equation}
Observed marking frequency alone cannot identify both error prevalence and
the probability of detecting and marking an error. Allowing erroneous or
confirmatory marks adds a term for $A=0$ and does not resolve the ambiguity.
This is a basic identification issue, not a new independence assumption.
It parallels the prevalence--labelling ambiguity in positive--unlabelled
learning \cite{bekker2020positive}; missing-data theory similarly makes
ignoring the observation mechanism conditional on explicit assumptions
\cite{rubin1976missing}. Interactive-learning interface research already
warns against interpreting unassigned examples as negative evidence
\cite{dudley2018review}.

An identification study can audit a randomly chosen subset of displayed but
unmarked regions using an independent reference or a separate inspection
protocol. Compare ignoring non-correction, treating it as acceptance, and
using calibrated soft evidence of acceptance. Exposure probabilities and
audit sampling must be recorded. Stated assumptions and direct validation
are preferable to silently turning every blank region into a pseudo-label.

\subsection{Does using selection improve assimilation?}
Compare ordinary supervised learning from addressed labels with a learner
that additionally models their selection, using the same correction stream
and fitting resources. Both receive the pixel addresses and labels; the
distinction is whether the selection mechanism informs the update.
Independent audit risk \eqref{eq:audit-improvement} measures the outcome
without recovering the source posterior. The excess-risk decomposition
\eqref{eq:assimilation-gap} explains why more informative input alone is
insufficient; BIFSeg illustrates the matched-correction comparison with a
current-image Dice endpoint \cite{wang2018bifseg}.

To test sequencing, replay the same multiset of corrections in different
orders at matched computational budgets, comparing delayed distortion and
errors in previously corrected regions. Curriculum and continual learning
motivate order, weighting, and replay interventions
\cite{bengio2009curriculum,delange2022continual}. Unlike image ordering that
also changes subsequent corrections \cite{torpey2026learning}, fixed-stream
replay separates these learner-side effects from acquisition.

Recorded corrections respond to particular displays: replay cannot reveal
what a human would supply under a changed policy. Repeats likewise do not
identify forgetting; delay, stale display, ambiguity, or annotation error
can also explain them.

\subsection{Does the interaction policy improve outcomes at matched effort?}
Test the closed-loop policy with actual interaction: changing views or updates
also changes what the teacher sees and supplies. Compare with strong
uncertainty-plus-diversity, BatchBALD where feasible, and core-set-style
baselines under common action, initialization, and fitting budgets. Record
view, checkpoint, support, labels, edits, empty responses, navigation, time,
and when corrections enter training. Participant and case allocation,
protocol order, and independent evaluation determine the causal interpretation.
Report paired or clustered uncertainty, the average performance trajectory,
and prespecified rare-structure failures. A favourable response score or
replay result motivates this test but does not establish its outcome.

\subsection{Extensions: teaching stages, exposure, and stopping}
Changes in teaching strategies \cite{cakmak2014eliciting,fiebrink2011evaluation}
and the greater search effort for small residual errors
\cite{benenson2019large} motivate stage-dependent models. RootPainter starts
with clear examples \cite{smith2022rootpainter}; progression through imbalance
correction to boundary refinement remains a hypothesis. Test stage features
beyond round, current error, class balance, expertise, fatigue, and display
delay, then evaluate any stage-dependent sampling or replay intervention.

Exploratory views and independent audits address blind spots outside a
learner's beliefs. Formal guarantees impose additional conditions.

Equivalence-class active learning groups hypotheses by a shared decision,
not necessarily by the evaluation-equivalence relation used here
\cite{golovin2010nearoptimal}. Greedy cardinality-budget guarantees require
adaptive monotonicity and submodularity; cost-to-cover results need stronger
conditions, including the corrected cover analysis of
\cite{golovin2011adaptive}. These are not guarantees for arbitrary neural
updates or human corrections. Overlapping noisy response distributions
require accumulated evidence, not elimination after one contradictory mark.

A low predicted gain or an empty correction does not certify small remaining
distortion. Confidence sequences support repeated monitoring under specified
sampling and estimand assumptions \cite{howard2021confidence}; applying them
requires representative observations and a declared risk for a changing
predictor. Neither an acquisition model's confidence nor a fixed training
budget supplies that evidence.

\section{Discussion and conclusion}
Three conclusions organize the evidence. First, response protocols affect
measured performance: RClicks changes benchmark difficulty, while Benenson
et al. link correction behaviour to search effort and completed-mask quality
\cite{antonov2024rclicks,benenson2019large}. Second, assimilation matters:
BIFSeg and Crayons compare ways of using supervision, rather than information
content alone \cite{wang2018bifseg,fails2003interactive}. Third, fewer
annotations do not establish less human effort, and completion quality does
not establish reusable-model performance. The Cellpose and microscopy SAM
comparisons illustrate why the cost unit and endpoint matter
\cite{pachitariu2022cellpose,archit2025microscopy}. Their heterogeneous outcomes
support these distinctions, not a ranking of systems.

The mathematical account explains when information produces expected log-loss
improvement and how an approximate update can lose it. Across losses,
independent evaluation can test predictive benefit without estimating exact
gain or identifying human intentions. The principal open question is whether
modelling selection and sequencing improves assimilation and, in actual
interaction, segmentation at matched effort beyond what addressed labels
and fitting resources explain.

The formal core holds the target fixed within a session. A changing
teacher interpretation, or one influenced by model exposure, would require
an additional state model; performative prediction is an adjacent
literature \cite{perdomo2020performative}. That extension is not needed merely
because RootPainter trains asynchronously.

\appendix
\section{Standard identities used in the comparison}
\label{app:derivations}
This appendix records the routine algebra underlying the main text. These
are established probability and information identities, not new results.
The application-specific issues are the choice of random objects, available
conditioning information, and learner update. All sums below use the support
of the corresponding distribution, with the zero-probability conventions
specified in Section~\ref{sec:distortion}. Finite-entropy or finite-loss
conditions are imposed wherever differences of quantities are taken.

\subsection{Log-loss risk, entropy, and distortion}
\label{app:risk-derivation}
For a fixed target, insert $P_f(y\mid x)$ into the logarithm:
\begin{align*}
 \CE(P_f,Q)
 &=-\sum_{x,y}\nu(x)P_f(y\mid x)\log_2 Q(y\mid x)\\
 &=-\sum_{x,y}\nu(x)P_f(y\mid x)\log_2 P_f(y\mid x)
   +\sum_{x,y}\nu(x)P_f(y\mid x)
        \log_2\frac{P_f(y\mid x)}{Q(y\mid x)}\\
 &=H_f(Y\mid X)+\sum_x\nu(x)
           \KL(P_f(\cdot\mid x)\Vert Q(\cdot\mid x))\\
 &=H_f(Y\mid X)+\tau(T_f,Q).
\end{align*}
The entropy term is independent of $Q$. For two predictors, possibly with
a random post-response predictor, it therefore cancels from the expected
risk difference in \eqref{eq:risk-reduction}. That cancellation holds for a
fixed $P_f$ and common evaluation weights, not for a changed target or
evaluation population. It introduces no relation to target entropy
$H(T_F)$.

\subsection{Predictive information and assimilation}
\label{app:predictive-information}
Condition throughout on $Z_i=z$ and use the fresh evaluation draw of
Section~\ref{sec:gain}. The Bayes predictors have expected log losses
$H(Y\mid X,z)$ and $H(Y\mid X,R_i,z)$. The expected target entropy floor
is unchanged by iterated expectation, so subtraction proves
\eqref{eq:predictive-information}.

For the actual predictors and excess risks defined in the main text, adding
and subtracting the respective Bayes log probabilities gives
\begin{align*}
 \E[\CE(P_F,Q_i)\mid z]&=H(Y\mid X,z)+a_i,\\
 \E[\CE(P_F,Q_{i+1})\mid z]&=H(Y\mid X,R_i,z)+a_{i+1}.
\end{align*}
Subtracting and cancelling the same expected target entropy floor proves
\eqref{eq:assimilation-gap}, under the stated finite-loss and conditioning
assumptions. Conditional data processing gives
$I(Y;R_i\mid X,z)\leq I(T_F;R_i\mid z)$: at evaluation-relevant addresses,
the fresh label depends on the response only through the target class.

\subsection{BALD as expected KL disagreement}
\label{app:bald-derivation}
Fix an address $x$ and observed memory $\cM$. Abbreviate
$w(P\mid\cM)$ by $w_P$ and $\widehat Q(y\mid x,\cM)$ by $\widehat Q_y$.
Using $\widehat Q_y=\sum_Pw_PP(y\mid x)$,
\begin{align*}
 \sum_Pw_P\KL(P(\cdot\mid x)\Vert\widehat Q(\cdot\mid x,\cM))
 &=\sum_{P,y}w_PP(y\mid x)\log_2 P(y\mid x)
   -\sum_y\widehat Q_y\log_2\widehat Q_y\\
 &=H(\widehat Q)-\sum_Pw_PH(P(\cdot\mid x)).
\end{align*}
Under the joint model $P(\mathsf P=P,Y=y\mid x,\cM)=w_PP(y\mid x)$,
this is $I(\mathsf P;Y\mid x,\cM)$. Identifying it with information about
the source target additionally requires the belief and target-relevance
conditions of Section~\ref{sec:bald-link}.

\subsection{Entropy, code length, and coding-model mismatch}
\label{app:code-derivation}
For completeness, if integer prefix lengths satisfy
$C=\sum_a2^{-\ell(a)}\leq1$, define $q(a)=2^{-\ell(a)}/C$ on the code
alphabet. Then
\[
 \sum_a p(a)\ell(a)=H(p)+\KL(p\Vert q)-\log_2 C\geq H(p).
\]
The choice $\ell(a)=\lceil-\log_2p(a)\rceil$ on positive-probability symbols
satisfies Kraft's inequality and gives an expected length less than $H(p)+1$.
The main text instead uses ideal probability-based lengths, for which
implementation overhead is separate.

At a fixed context $z$, write $\rho_i(r\mid z)$ for the true response mass.
Adding and subtracting its log probability gives
\begin{align*}
 \E[\ell_i\mid Z_i=z]
 &=-\sum_r\rho_i(r\mid z)\log_2\widehat\rho_i(r\mid z)\\
 &=-\sum_r\rho_i(r\mid z)\log_2\rho_i(r\mid z)
   +\sum_r\rho_i(r\mid z)
       \log_2\frac{\rho_i(r\mid z)}{\widehat\rho_i(r\mid z)}\\
 &=H(R_i\mid Z_i=z)
   +\KL(\rho_i(\cdot\mid z)\Vert\widehat\rho_i(\cdot\mid z)).
\end{align*}
Averaging over $Z_i$ proves \eqref{eq:code-mismatch}. The other step in
Proposition~\ref{prop:coding} is
\[
 I(T_F;R_i\mid Z_i)
 =H(R_i\mid Z_i)-H(R_i\mid T_F,Z_i)
 \leq H(R_i\mid Z_i).
\]
Summing over rounds proves \eqref{eq:code-bound}. Nonnegativity of the
last conditional entropy uses discrete responses; it is not a general
property of differential entropy.

\subsection{Posterior-entropy averaging and telescoping gains}
\label{app:gain-derivation}
For a discrete memory value $m$, let $p_m(t)=P(T_F=t\mid M_i=m)$. Then
\begin{align*}
 \E_{M_i}[H(p_{M_i})]
 &=-\sum_m P(M_i=m)\sum_t P(T_F=t\mid M_i=m)
                         \log_2P(T_F=t\mid M_i=m)\\
 &=-\sum_{m,t}P(T_F=t,M_i=m)\log_2P(T_F=t\mid M_i=m)\\
 &=H(T_F\mid M_i).
\end{align*}
For non-discrete memory the outer sum is an integral with respect to its
distribution; the target entropy remains discrete. No independence between
target and memory is assumed.

Write $h_i(m)=H(T_F\mid M_i=m)$. Along an actual history, the hypothetical
record with response $r=R_i$ is the actual $M_i$, so
\begin{align*}
 \sum_{i=1}^n\Gain_i(R_i;Z_i)
 &=\sum_{i=1}^n\bigl[h_{i-1}(M_{i-1})-h_i(M_i)\bigr]\\
 &=H(T_F)-h_n(M_n).
\end{align*}
Taking the joint source/history expectation and using the preceding identity
gives \eqref{eq:gain-telescope}. A fixed-target expectation instead averages
with a different distribution and does not in general yield this mutual
information.

The memory decomposition \eqref{eq:full-information} remains in the main
text because it determines which protocol choices must be accounted for.
Under \eqref{eq:no-fresh-context}, its context terms vanish and
\[
 \E\sum_{i=1}^n\Gain_i(R_i;Z_i)
 =\sum_{i=1}^nI(T_F;R_i\mid Z_i)
 \leq\ell_{\mathrm{PT}}(n),
\]
which proves Proposition~\ref{prop:full-gain}. This uses nested memory and
the stated protocol condition, not interchangeability of $M_n$ and
$\mathbf R_n$.

\subsection{Support and label chain rules}
\label{app:support-derivation}
At each fixed context, the probability product rule gives
$\widehat\rho_i(\xi,h\mid z)=\widehat\rho_i(\xi\mid z)
\widehat\rho_i(h\mid\xi,z)$. Taking negative logarithms, summing over
rounds, and taking expectations gives \eqref{eq:support-label-code}.
For a shared ordering $\xi=\{x_1,\ldots,x_m\}$, the second factor is
\[
 \widehat\rho_i(h\mid\xi,z)
 =\prod_{j=1}^m
   \widehat\rho_i\bigl(h(x_j)\mid\xi,z,h(x_1),\ldots,h(x_{j-1})\bigr).
\]
The factors denote the corresponding conditional label probabilities derived
from the same response model. This gives one label term per marked location,
not independent observations or one optimiser update per term.

The information identity follows by adding and subtracting the entropy
after support is known:
\begin{align*}
 I(T_F;\Xi_i,H_i\mid Z_i)
 &=H(T_F\mid Z_i)-H(T_F\mid Z_i,\Xi_i,H_i)\\
 &=\bigl[H(T_F\mid Z_i)-H(T_F\mid Z_i,\Xi_i)\bigr]\\
 &\quad+\bigl[H(T_F\mid Z_i,\Xi_i)-H(T_F\mid Z_i,\Xi_i,H_i)\bigr]\\
 &=I(T_F;\Xi_i\mid Z_i)+I(T_F;H_i\mid\Xi_i,Z_i).
\end{align*}
Neither the coding factorization nor this information decomposition assumes
independence of support and labels.

\section{The parametric MDL benchmark and its qualifications}
\label{app:mdl}
Unlike the algebra in \ref{app:derivations}, this asymptotic benchmark
requires substantive statistical assumptions. We retain the construction to
clarify its interpretation, not as a theorem about adaptive human correction
streams. For broader treatments see \cite{grunwald2007mdl,grunwald2020revisited}.

Hold the discrete observation alphabet or finite-precision recording scheme
fixed as $N$ varies. All likelihoods below are probability masses of the recorded
data, not density values; parameters may still be continuous. The regularity
assumptions apply to this recorded-data model.

Consider a \emph{fixed} regular $k$-parameter model for well-specified i.i.d.
data, an interior true parameter with nonsingular Fisher information, and
the usual smoothness and concentration conditions. Write $\widehat\theta_N$
for its maximum-likelihood fit. A suitable parameter grid at scale
$N^{-1/2}$ in each coordinate uses approximately $(k/2)\log_2 N$ bits on a
fixed bounded parameter region. Under regularity, rounding the fit to that
grid changes the data log loss by a bounded-order amount. A suitable
two-part code consequently has the benchmark expansion
\begin{equation}
 \ell_{\mathrm{two}}(D_N)
 =-\log_2 p_{\widehat\theta_N}(D_N)
   +\frac{k}{2}\log_2 N+O_P(1).
 \label{eq:mdl-two-benchmark}
\end{equation}
Here $O_P(1)$ is bounded in probability, not a uniform bound for every possible
dataset. The precision, model family, and code construction are essential.

For a smooth positive prior density $\pi$ near the true parameter, the
Bayesian mixture code has an exact prequential factorization:
\begin{equation}
 q(D_N)=\int p_\theta(D_N)\pi(\theta)\,d\theta
       =\prod_{t=1}^N q(d_t\mid D_{t-1}).
 \label{eq:mixture-code}
\end{equation}
Under regular Laplace-approximation conditions, its negative log probability
has the same leading likelihood and $(k/2)\log_2 N$ terms as
\eqref{eq:mdl-two-benchmark}. If the remainder also has uniformly bounded
absolute expectation, taking expectations yields
\begin{equation}
 \E[\ell_{\mathrm{preq}}(D_N)]
 =\E[-\log_2 p_{\widehat\theta_N}(D_N)]
  +\frac{k}{2}\log_2 N+O(1).
 \label{eq:mdl-preq-benchmark}
\end{equation}
Boundedness in probability alone would not
justify that expected expansion.

This gives a concrete sufficient coding construction; it is not a guarantee
for an arbitrary plug-in training sequence. For example, unsmoothed Bernoulli
maximum likelihood after the first observation assigns probability zero to
the opposite outcome and can produce infinite subsequent code length.

The term describes redundancy relative to a retrospectively fitted likelihood
within the stated family. It is neither all the communication required nor
a universal lower bound on teacher effort. If model dimension is selected
from varying $\Theta_k$, describing the family index has a further cost.
For a response model, $k$ would be its identifiable statistical dimension,
not automatically the number of segmentation-network parameters or target
bits. Spatial dependence, adaptive views, misspecification, and changing
teacher behaviour all require separate analysis. In particular, neither the
number of rounds $n$ nor the number of marks $W_n$ can automatically replace
the i.i.d. sample size $N$ in \eqref{eq:mdl-preq-benchmark}.

\section{Literature-selection and search record}
\label{app:search-record}

\subsection{Question, design, and scope}
The review asks when corrective spatial feedback is informative, when it
becomes useful learning, and what an improvement per human effort establishes.
It is a critical narrative review with purposive source selection, not a
systematic review or meta-analysis. The anchor corpus combines established
methods, prior reviews, representative corrective-segmentation systems, and
statistical foundations; it was not prospectively sampled under a registered
search. A targeted search on \textbf{14 October 2026} checked recent
counterexamples and missing mechanisms. It does not establish comprehensive
coverage of 2024--2026.

Discovery used a general web-search service, year terms, and targeted
publisher/proceedings searches, followed by primary-source retrieval. Ranked
results change over time; no total-hit or PRISMA-style count is claimed.
News, aggregators, and social media supplied discovery leads only. Year terms
are not verified publication-date filters: dates and publication status were
checked against primary records, with preprints identified as such.

Sources were included when they clarified who chooses views, response
locations, labels, or correction extent; actual versus simulated responses
and their observation models; prompt-state updates, image-specific adaptation,
or reusable learning; or the endpoint, denominator, and design underlying an
efficiency claim. Theory and nonsegmentation studies supply mechanisms or
assumptions, not direct evidence of human segmentation efficacy. General
architectural advances without a relevant interaction, selection, assimilation,
or evaluation mechanism were excluded; exhaustive coverage of SAM variants
was outside scope. Empirical claims were deferred when the primary protocol
could not be inspected. Reviews positioned the synthesis but did not replace
primary evidence of individual systems' performance.

\subsection{Exact queries}
The following strings preserve the discovery and retrieval queries, including
quotation marks, year terms, site restrictions, and exclusions. The targeted
checks were not an exhaustive forward/backward citation search.

\begingroup
\small\raggedright
\paragraph{Discovery}
\begin{description}
\setlength{\itemsep}{1pt}\setlength{\parsep}{0pt}
\item[Q1] \texttt{2024 2025 2026 interactive segmentation human corrective learning annotation selection retraining}
\item[Q2] \texttt{2024 2025 2026 interactive segmentation human click model user study correction behavior}
\item[Q3] \texttt{2024 2025 2026 medical segmentation annotation effort human in the loop correction active learning}
\item[Q4] \texttt{"interactive segmentation" "human" "click" "2025" site:openaccess.thecvf.com}
\item[Q5] \texttt{"corrective" "segmentation" "2025" "learning" site:arxiv.org}
\item[Q6] \texttt{"segmentation" "human" "annotation time" "2026" "learning" site:pubs.rsna.org}
\end{description}

\paragraph{Candidate metadata and retrieval}
\begin{description}
\setlength{\itemsep}{1pt}\setlength{\parsep}{0pt}
\item[F1] \texttt{"XpertLoop" human annotation learning 2026}
\item[F2] \texttt{"SPA" "Efficient User-Preference Alignment" arxiv}
\item[F3] \texttt{"SegAgent" "Imitating Human" arxiv}
\item[F4] \texttt{"Segment anything in medical images" "annotation study" "10"}
\item[F5] \texttt{"XpertLoop" "doi"}
\item[F6] \texttt{"Towards Clinician-Preferred Segmentation" "10.1007"}
\item[F7] \texttt{"XpertLoop" "Jaiswal" "Radiology" doi}
\end{description}

\paragraph{Mechanism and publication checks}
\begin{description}
\setlength{\itemsep}{1pt}\setlength{\parsep}{0pt}
\item[T1] \texttt{"RClicks" "teacher" "learning"}
\item[T2] \texttt{"Iterative Teacher-Aware Learning" segmentation human}
\item[T3] \texttt{"RootPainter" "click" "model" human correction selection}
\item[T4] \texttt{"Cellpose" "human" "correction" "selection" learning}
\item[T5] \texttt{"RClicks" "learning" "human" "training" 2025 2026}
\item[T6] \texttt{"Iterative Teacher-Aware Learning" "segmentation" -site:researchgate.net}
\item[T7] \texttt{"RClicks" "retraining" segmentation}
\item[T8] \texttt{"You Point, I Learn" DOI 2025}
\item[T9] \texttt{"You Point, I Learn" "ICLR 2026" Xu Liang}
\item[T10] \texttt{"MultiVerSeg" "time" "human" ICCV 2025}
\item[T11] \texttt{MultiverSeg Scalable Interactive Segmentation Biomedical Imaging Datasets In Context Guidance DOI authors}
\item[T12] \texttt{"MultiverSeg" site:arxiv.org}
\item[T13] \texttt{"Active Learning for Efficient Annotation of Surgical Videos with Weak Supervision" journal DOI}
\item[T14] \texttt{"Explainable Human-in-the-Loop Segmentation via Critic Feedback Signals" publication DOI}
\item[T15] \texttt{"Learning What To Ask For When" Torpey}
\item[T16] \texttt{"Learning What To Ask For When" arxiv}
\item[T17] \texttt{"Torpey" "Image Ordering" segmentation pdf}
\item[T18] \texttt{"Learning What To Ask For When" "simulat"}
\item[T19] \texttt{"Learning What To Ask For When" "context" site:openaccess.thecvf.com/content/CVPR2026W}
\end{description}

\paragraph{XpertLoop retrieval checks}
\begin{description}
\setlength{\itemsep}{1pt}\setlength{\parsep}{0pt}
\item[X1] \texttt{XpertLoop Jaiswal 2026 annotation human expert segmentation}
\item[X2] \texttt{"XpertLoop"}
\item[X3] \texttt{"XpertLoop" "radiologist"}
\item[X4] \texttt{"XpertLoop" "Jaiswal"}
\item[X5] \texttt{"Jaiswal" "human-guided" annotation radiology 2026}
\item[X6] \texttt{site:pubs.rsna.org "Jaiswal" "annotation"}
\item[X7] \texttt{"expert-guided annotation" "doi"}
\item[X8] \texttt{"XpertLoop" segmentation -site:linkedin.com -site:xpertloop.ai}
\item[X9] \texttt{"10.1148/ryai.250200"}
\item[X10] \texttt{"Impact on Cost and Expert Time"}
\end{description}
\endgroup

\subsection{Retrieval paths and screening decisions}
Mechanism-focused discovery retrieved ExiL, whose references led to
\emph{You Point, I Learn} and MultiverSeg; their primary texts and proceedings
were checked. MultiverSeg-related discovery led to Torpey et al., verified
against the indexed CVF record and proceedings PDF. XpertLoop title/DOI
checks led to the RSNA article and PubMed record. Dendukuri's title search
led to the preprint, \href{https://liaoguiqiu.github.io/news/}{author announcement},
and \href{https://cars-int.org/wp-content/uploads/2026/06/CARS2026_program_single_pages-003.pdf}{official IPCAI long-abstract programme};
Shaeri's led to the preprint and an
\href{https://rtwoo.github.io/publications/}{author publication-status page}.
These paths establish relevant contrasts, not citation-network
saturation.

\paragraph{Included candidates with primary text inspected}
MedSAM~\cite{ma2024medsam} was included for measured annotation-pipeline
time; SPA~\cite{zhu2025spa} for preference observations and tested
cross-image preference-state transfer; and HiTTA~\cite{hu2024hitta} for
cross-image assimilation under simulated preferences. SegAgent~\cite{zhu2025segagent}
was retained briefly as a boundary case separating constructed annotation
trajectories from observed human trajectories. Dendukuri et
al.~\cite{dendukuri2026active}, Shaeri et al.~\cite{shaeri2025explainable},
and ExiL~\cite{tavangar2026exil} provide adjacent evidence of timed corrective
learning under distinct protocols. XpertLoop~\cite{jaiswal2026xpertloop}
enters Table~\ref{tab:interaction-evidence} as a direct correction/retraining
and expert-time study. MultiverSeg~\cite{wong2025multiverseg} supplies an
in-context reuse precedent; \emph{You Point, I Learn}~\cite{xu2026youpoint}
an online parameter-assimilation precedent; and Torpey et
al.~\cite{torpey2026learning} simulated sequencing evidence. Their inspected
versions, locators, observation units, and inferential limits are recorded
in \ref{app:evidence-appraisal}.

\paragraph{Other screened hits}
\begin{description}
\item[PRISM] \emph{A Promptable and Robust Interactive Segmentation Model
with Visual Prompts} (\href{https://pmc.ncbi.nlm.nih.gov/articles/PMC12128912/}{PMCID: PMC12128912}).
The primary indexed description concerns simulated corrective prompts and
iterative mask refinement. Not selected for detailed appraisal because
representative fixed-model refinement methods were already included;
cataloguing further prompt architectures would broaden the scope.
\item[CASC-AI] \emph{Consensus-aware Self-corrective AI Agents for Noise Cell
Segmentation} (\href{https://arxiv.org/abs/2502.07302}{arXiv:2502.07302}).
The primary abstract concerns offline noisy-label consensus/weighting, not
live selection or sequencing of human corrections. Not selected: annotator
noise was already addressed by sources matched to the relevant assumptions.
\item[Predictive-corrective brain MRI] \emph{Rethinking Convergence in Deep
Learning: The Predictive-Corrective Paradigm for Anatomy-Informed Brain MRI
Segmentation} (\href{https://arxiv.org/abs/2510.15439}{arXiv:2510.15439}).
The primary abstract uses ``corrective'' for a residual network, not human
feedback. Excluded on interaction-scope grounds.
\item[RAIL] \emph{Region-Aware Instructive Learning for Semi-Supervised Tooth
Segmentation in CBCT} (\href{https://arxiv.org/abs/2505.03538}{arXiv:2505.03538}).
In the primary abstract, teacher/student denotes neural training components,
not human example selection. Excluded from the human-interaction synthesis.
\item[Segment Anything for Microscopy] The 2025 study~\cite{archit2025microscopy}
was already included in the detailed evidence matrix; the discovery hit
does not constitute a separate study. Its user study is relevant to effort.
\item[McSweeney et al.] \emph{Transfer learning for data-efficient abdominal
muscle segmentation with convolutional neural networks}
(\href{https://doi.org/10.1002/mp.15533}{doi:10.1002/mp.15533}, 2022).
The primary paper, Sections 2.1--2.4, uses upfront expert annotations,
offline transfer learning on random training subsets, and a ten-radiographer
interobserver benchmark. It evaluates no interactive correction, adaptive
selection, or annotation-time intervention and was not selected for corrective
feedback comparisons.
\end{description}

\paragraph{Coverage limits}
The included sources cover corrective retraining, in-context reuse, response
simulation, expert-guided adaptation, and ordering. They support neither
exhaustive coverage nor a claim that selection/history has no precedents.
Primary full texts support the sequencing and adaptation comparisons;
XpertLoop was appraised from the publisher's full text.
No exhaustive database export or complete citation-network sweep was
performed. Independent appraisal of consequential evidence classifications
remains desirable. These are limits on coverage and confidence, not evidence
against individual studies.

\section{Study selection and qualitative appraisal}
\label{app:evidence-appraisal}

\subsection{Comparison and extraction rule}
The studies in Table~\ref{tab:interaction-evidence} were selected to illustrate
distinct links between selection, assimilation, and effort: reusable
classification, sparse corrective learning, longitudinal clinical correction,
current-image adaptation, completed-mask retraining, workflow integration,
human-response simulation, prompting plus reuse, cost-aware acquisition, and
clinically timed iterative retraining. This is purposive coverage of
mechanisms, not a representative sample of studies or applications.

Comparisons were selected for those links, not to maximize effects or rank
systems. The appraisal records the unit, participants, and horizon;
intervention, comparator, and controls; endpoint, cost, and uncertainty unit;
and supported inference. Alternative conditions qualify the selected
comparisons but do not constitute an exhaustive sensitivity audit. NR means
not reported for the selected comparison, not an observed null result.
Uncontrolled order and bundled interventions are design limitations distinct
from missing reporting. SD, SE, and confidence intervals are not
interchangeable, and none automatically quantifies variability across human
users. Numerical outcomes for the main comparisons remain in
Table~\ref{tab:interaction-evidence}; the records below supply source locators,
selection rationale, and additional qualifications. Page numbers refer to
printed pagination unless explicitly identified as PDF pages.

\subsection{Appraisal of the main comparisons}

\paragraph{Crayons~\cite{fails2003interactive}}
\emph{Source:} p.~44, Tables 1--2 and Evaluation.
Ten participants in two counterbalanced order groups performed four
visual-classification tasks. The endpoint was elapsed time to a reusable
classifier exceeding 97.5\% agreement with a reference classifier. The skin
task contrasts stronger generalization (CWSS) with faster individual fits
(MSSS), illustrating that fitting speed and interactive efficiency differ.
Intervals/tests are NR. Differences are smaller for simpler tasks; paper
cards slightly favour MSSS in one order. Reference-classifier agreement is
not a population-risk test.

\paragraph{RootPainter~\cite{smith2022rootpainter}}
\emph{Source:} pp.~780--782, Tables 2--3 and Fig.~4.
Two author-users trained on three plant-image datasets, with corrective
annotation always preceding duration-matched dense annotation. Held-out
downstream measurements assess the final reusable predictors. The selected
comparison directly matches human time; it does not establish superiority.
The reported means and SEs aggregate six user--dataset runs, and the
mixed-effects strategy test is nonsignificant at the stated 0.05 level.
Dense annotation wins for one nodule user; root results are close. SE is
not an effect confidence interval, and order was uncontrolled.

\paragraph{RootPainter3D~\cite{smith2022rootpainter3d}}
\emph{Source:} pp.~463, 466--467, Fig.~5 and Table~1.
One physician contoured 933 CT hearts with cumulative training; twenty were
also contoured in Eclipse. The final-ten comparison demonstrates late-session
completion savings, including inspection/correction but excluding inactivity
of at least twenty seconds. Timing confidence intervals are NR. Early
assisted cases are slower, and final-case times omit accumulated
initialization effort. Case order, user experience, and training evolve
together. Agreement between corrected outputs and manual contours does not
establish independent accuracy of every uncorrected prediction.

\paragraph{BIFSeg~\cite{wang2018bifseg}}
\emph{Source:} final TMI 2018 article, pp.~1567, 1571--1572, Tables V--VI.
Two clinical users evaluated fetal and brain-tumour MRI refinement. The
selected brain comparison fixes initial predictions and one round of
scribbles, comparing BIFSeg with CRF. Shared supervision isolates differences
in assimilation more directly than changing a query policy. The tables also
compare unsupervised refinement and unweighted fine-tuning: assistance works
without extra scribbles, but supervised refinement improves further.
Parameters adapt to the current image, not subsequent cases. The table's
$\pm$ dispersion is not explicitly identified as SD or SE.

\paragraph{Cellpose 2.0~\cite{pachitariu2022cellpose}}
\emph{Source:} pp.~1639--1640, Fig.~5.
One annotator supplied five training and three test images in each of two
categories. The online/pretrained versus offline comparison connects
completed-mask retraining to later-image prediction; cost-ratio intervals
are NR. The factorial comparison also includes online/from-scratch models,
which require 370 and 528 drawn objects, showing a material contribution
from initialization. Repeated test annotation measures within-annotator
consistency, not generalization across annotators. Drawn-object counts do
not establish equal session duration.

\paragraph{MONAI Label~\cite{diazpinto2024monai}}
\emph{Source:} PDF pp.~7--8, Section 5.1, Tables 1--2.
The spleen-CT demonstration adds corrected volumes across annotation stages.
It was selected as workflow evidence, not an acquisition-policy experiment.
Manual paintbrush, contour, and scribble baselines change the assistance
comparison; training time is reported separately. Uncertainty intervals and
user-population variability are NR. Six validation volumes in the prose
versus nine in Table~2, together with inconsistent timing totals, prevent
precise reconstruction of a cost-effect estimate. These are reporting
ambiguities, not measured evidence against efficiency.

\paragraph{RClicks~\cite{antonov2024rclicks}}
\emph{Source:} PDF pp.~7--9, Section 5 and Table~4; collection in Section 3.
Real clicks train a response-location model; simulations compare clicking
strategies across segmentation models using clicks to 90\% IoU, capped at
twenty. This was selected to show that response assumptions alter benchmark
difficulty. SD concerns sampled click-group variability; timing uncertainty
is inapplicable because human time was not measured. Effects vary by model
and dataset. Selected model-error displays supply later-click tasks, rather
than unconstrained longitudinal human sessions. This comparison does not
improve a reusable learner.

\paragraph{Microscopy SAM~\cite{archit2025microscopy}}
\emph{Source:} p.~586, User study 1; p.~588, Fig.~6a.
Five organoid annotators compared fine-tuned $\mu$SAM and CellPose. The
comparison distinguishes annotation completion from subsequent reusable-model
outcomes; it does not prove a significant rank reversal. Test mSA dispersion
is SD across annotators. The alternative manual condition averages
$6.7\pm0.9$ seconds/object and mSA $0.49\pm0.07$. Initialization and
downstream training belong to the pipeline, whereas per-object annotation
rates omit complete training-plus-human cost.

\paragraph{CEREALS~\cite{mackowiak2018cereals}}
\emph{Source:} PDF pp.~8--10, Section 4, Figs.~4--5.
Five Cityscapes simulations use a label oracle and recorded polygon-click
cost. The selected $128\times128$-region comparison isolates learned
cost-aware versus entropy-only acquisition under that simulation, not human
time. Threshold-cost intervals are NR. Oracle-cost acquisition reaches the
same target at 14.68\% of annotation cost, exposing the gap introduced by
cost prediction. The target is 95\% of full-data mIoU 0.605, not 95\%
absolute mIoU. Region size and the recorded-click cost model condition the
comparison.

\paragraph{XpertLoop~\cite{jaiswal2026xpertloop}}
\emph{Source:} published RSNA primary text, pp.~2, 5, 8--11 and Fig.~5;
annotation-time and cost analyses. The study uses 1,941 CT/MRI scans from
ten datasets and 57 nnU-Net models. It retrains nnU-Net from scratch on
accumulated corrected masks, comparing random and size--entropy selection.
It was selected because live expert timing and reusable-model training
coexist. One radiologist per timed task annotated different randomly selected
case subsets manually or corrected optimized-model predictions. Maximum
kidney/tumour savings must not be generalized across tasks. Modelled tumour
cost uncertainty includes no saving and derives from assumed wage/time
distributions, not multiple clinicians. Infrastructure and development costs
are excluded. This is neither a matched end-to-end training-budget comparison
nor an isolated benefit of modelling human selection.
The acquisition analysis (pp.~3, 7, 10--11, Tables 2--3 and Discussion)
reports fewer loops to strong performance for size--entropy than random
selection in the reported organ tasks; plain entropy can underperform in
the ablations, and suitability depends on the dataset. Size--entropy adds
inference over unlabelled cases. Mean test-set Dice improvement below 2\%
determines stopping, so that set informs the procedure rather than providing
an untouched evaluation. Loop counts alone do not compare total human-plus-
computational cost at independent evaluation.

\subsection{Adjacent cases and source checks}
These sources clarify nearby mechanisms without entering a pooled numerical
comparison. Publication status and available evidence are stated separately;
absence of measured human time is not treated as a null benefit.

\paragraph{Large-scale interactive segmentation~\cite{benenson2019large}}
\emph{Source:} CVPR 2019 work, inspected arXiv PDF pp.~5--8,
Sections 4.1--4.2 and 5.1--5.4, Figs.~7--9.
A pool of 100 full-time annotators supplies up to four clicks per round in
three rounds separated by days. Zero-click judgments take about four
seconds; four-click rounds are faster than three-click rounds. Mean round
times increase from 10.8 to 11.7 to 11.9 seconds, consistent with smaller
errors taking longer to locate. The proposed explanation of the three-
versus-four-click pattern is a hypothesis, not an identified causal effect;
timing-effect confidence intervals are NR. A ranker uses click counts,
round number, mask change, and click-to-mask distances to predict completed-mask
quality (Sections 4.2 and 5.4, Fig.~8). These results support accounting for
inspection, changing correction behaviour, and history-dependent quality
signals within the protocol, not universal teaching stages. A separate
downstream experiment trains on completed masks (Section 5.3) without testing
whether using the ranker improves reusable learning. The headline speed
comparison uses 34 measured seconds per corrected mask against 108 estimated
seconds for historical COCO polygons, scaling historical speed linearly by
vertex count (Section 5.1); it is not a matched-user timing trial.

\paragraph{MedSAM~\cite{ma2024medsam}}
\emph{Source:} final 2024 \emph{Nature Communications} article; Results,
``MedSAM can improve the annotation efficiency''; Methods, ``Human annotation
study''; Fig.~5b. Two radiologists annotated ten adrenal-tumour
cases manually and, after a one-week interval, with model assistance.
Assisted time includes initial markers, inference, and manual refinement.
This is an actual human-time comparison of annotation pipelines; the fixed
order and small participant sample limit causal and population claims. It
does not demonstrate reusable learning from those corrections.

\paragraph{SPA~\cite{zhu2025spa}}
\emph{Source:} final ICCV 2025 paper, Sections 3.1--3.5 and 4.3.2,
pp.~23735, 23737--23738;
\href{https://openaccess.thecvf.com/content/ICCV2025/supplemental/Zhu_SPA_Efficient_User-Preference_ICCV_2025_supplemental.pdf}{official supplement}
for protocol details.
Multiple-choice mask selection updates a latent preference distribution;
training simulates feedback using multi-rater masks. Section 3.4 initializes
preference state per image, while Section 4.3.2 also transfers user-adapted
GMM parameters across images (kidney Dice 91.68\% to 92.43\%). A separate
study with five medical participants and 100 fundus images reports shorter
interactions than MedSAM. That user study does not isolate transfer benefits.
Mask choice is a different response unit from spatial correction: reusable
preference state is not a learned likelihood of natural support choice.
The latent-Gaussian/i.i.d. convergence claim is not a guarantee for general
spatial correction behaviour.

\paragraph{HiTTA~\cite{hu2024hitta}}
\emph{Source:} arXiv:2405.08270v1, Sections 2.1, 2.3, 3.1--3.4 and
Table~2; no peer-reviewed record was verified in the search.
Corrected masks update a model used on subsequent images. Existing
ophthalmologists' masks assigned to different domains simulate preferences;
these are not live timed corrections. Section 3.2 (p.~6) evaluates the
better prediction from the original and preference heads. Domain-preference
DSC is 91.14\% versus 87.21\% without human feedback, with no uncertainty
intervals. The result supports adaptation under this simulated protocol,
not demonstrated human-effort efficiency or an independently validated,
deployable head-selection rule.

\paragraph{SegAgent~\cite{zhu2025segagent}}
\emph{Source:} final CVPR 2025 paper, Sections 3 and 4.1 (p.~3689), and
the trajectory-generation algorithm.
An automated click simulator generates training trajectories from
reference-mask errors; these are not recorded human trajectories. Successful
imitation establishes agent performance under that generator, not realism
of human-response probabilities or improvement in reusable segmentation
learning at matched human effort. This is the boundary case for which the
study was included.

\paragraph{Surgical-video annotation~\cite{dendukuri2026active}}
\emph{Source:} arXiv:2607.13237v1, Sections 2.2, 3.2--3.2.2, Tables 1--3.
The inspected full paper remains a preprint; a related IPCAI 2026
long-abstract acceptance was verified separately against the author
announcement and official programme, not treated as full-paper publication.
Two trained annotators supply corrections for later training; acquisition
is random with class-balancing weights. AL-10 averages 3.56 minutes/video
versus 6.34 manually, but SAM2 assistance takes 1.60. CorLoc gains over
upfront supervision depend on budget and loss weighting; AL-2 can
underperform. Video counts differ between prose and Table~1. Different
annotation aids enter the comparison, precluding attribution to selection
modelling alone or a general 50\% effort-saving claim.

\paragraph{Critic feedback~\cite{shaeri2025explainable}}
\emph{Source:} arXiv:2510.09945v1, Sections 3.2.4, 3.3, 4.4 and 5.1--5.4.
The author page lists WACV submission, not verified acceptance.
Twelve participants corrected twenty images each with randomized task order.
Corrected-label fitting and similarity-based propagation accompany the
interface. Reported times are 24 seconds/image versus 54 for click refinement
and 95 for pixel annotation; timing intervals are NR. Editing, feedback,
propagation, and fitting change together. Section 5 describes variability as
confidence intervals, whereas Table~1 labels it seed SD; seed variability
does not quantify a human-time effect. This is a qualified cross-image
feedback precedent, not causal identification or an isolated natural-selection
model.

\paragraph{ExiL~\cite{tavangar2026exil}}
\emph{Source:} accepted MICCAI 2026 full text, LNCS 16884, Sections 2.2--2.5
and 3, especially Section 2.5 (pp.~6--7) and Table~2. The official proceedings
page had not supplied a DOI or final pagination at the check.
Synthetic brush trajectories teach refinement; expert-approved masks and
intermediate correction states update the deployed model. One expert timed
360 frames from three volunteers with randomized frame order and common
stopping/acceptance guidance. This is live within-annotator feasibility
evidence, with limited transfer across users. The synthetic autofill
comparison restricts SAM to one refinement; the timed SAM-assisted workflow
instead reports 2.9 autofill cycles/frame. A completion trajectory is
modelled, not a likelihood of natural human support choices.

\paragraph{MultiverSeg~\cite{wong2025multiverseg}}
\emph{Source:} arXiv:2412.15058v2 of the ICCV 2025 work, Sections 3 and 5--6;
publication metadata checked against the published record.
Completed image--mask pairs enter the context for subsequent images without
within-session weight retraining. Simulated clicks and scribbles assess how
earlier completed cases aid later ones. This is evidence of reusable state,
not measured human-time savings: cross-image learning need not entail
parameter updating.

\paragraph{You Point, I Learn~\cite{xu2026youpoint}}
\emph{Source:} arXiv:2503.06717v4 (20 April 2026), marked as published at
ICLR 2026 and checked against the official proceedings; Sections 2.1, 2.4,
3.2 and Appendices A.2, A.4.
Mid- and post-interaction parameter updates reuse corrections across images
under distribution shift. Inference clicks are simulated in the largest
reference-mask error component; live human time is not measured and natural
correction preferences are not estimated. Appendix A.4 shows that
post-interaction gains can diminish by the last click, so benefits depend
on evaluation horizon and click budget.

\paragraph{Image ordering~\cite{torpey2026learning}}
\emph{Source:} final CVPR Workshops (PHAROS-AIF-MIH) 2026 paper,
Sections 4.2--4.5, 5.2 and 6.4, pp.~6420--6425, Tables 1--2.
Ten-image subsets are evaluated under 100 random orders with fixed
MultiverSeg weights. Completed predictions or reference masks enter later
context; deterministic clicks follow reference-mask errors. Across ten
datasets with predicted-mask context, the sampled best--worst Dice range is
6.07 percentage points larger at five than twenty clicks (95\% CI
5.62--6.55). IQR comparisons also show order sensitivity. Confidence intervals
bootstrap image subsets with hierarchical aggregation, not users.
Uncertainty ordering averaged across starts has mixed effects versus random,
with most intervals crossing zero; selecting the best of ten starts
retrospectively is more favourable. This separates order sensitivity from a
reliable acquisition rule. Click ceilings condition cost comparisons; neither
count reductions nor oracle-start effects establish human-time savings.
Order changes subsequent corrections, so this does not replay a fixed
correction multiset.

\paragraph{Prior-review positioning~\cite{marinov2024review}}
\emph{Source:} final TPAMI 2024 review, pp.~11010--11013,
Sections V-A3, V-D and VI-B1--B3.
Beyond training/application categories, Marinov et al. appraise absent or
weak baselines, fragmented comparisons, narrow accuracy metrics, and scarce
user-centred measures. Section VI-B2 discusses the difficulty of simulating
expert annotation styles and recommends diverse simulated users; VI-B3
distinguishes annotation efficiency from interaction efficacy. These are
established evaluation concerns. This review's additional role is to
formalize assumptions connecting selected responses, learner updates, and
declared effort. This entry positions a review; it is not primary effect
evidence.

\paragraph{Synthesis boundary}
These contrasts support protocol-specific conclusions about fitting,
completion, response simulation, and cost-aware acquisition. No aggregate
effect is calculated across endpoints, uncertainty units, or cost measures.
The absence of one complete selection-model-to-live-efficiency demonstration
within this set is not evidence that none exists elsewhere.

\bibliographystyle{apalike}
\bibliography{survey_refs}

@article{ma2024medsam,
  author = {Ma, Jun and He, Yuting and Li, Feifei and Han, Lin and You, Chenyu and Wang, Bo},
  title = {Segment anything in medical images},
  journal = {Nature Communications},
  volume = {15},
  pages = {654},
  year = {2024},
  doi = {10.1038/s41467-024-44824-z}
}

@inproceedings{zhu2025spa,
  author = {Zhu, Jiayuan and Wu, Junde and Ouyang, Cheng and Kamnitsas, Konstantinos and Noble, J. Alison},
  title = {{SPA}: Efficient User-Preference Alignment against Uncertainty in Medical Image Segmentation},
  booktitle = {Proceedings of the IEEE/CVF International Conference on Computer Vision},
  pages = {23731--23740},
  year = {2025},
  url = {https://openaccess.thecvf.com/content/ICCV2025/html/Zhu_SPA_Efficient_User-Preference_Alignment_against_Uncertainty_in_Medical_Image_Segmentation_ICCV_2025_paper.html}
}

@misc{hu2024hitta,
  author = {Hu, Shishuai and Liao, Zehui and Liu, Zeyou and Xia, Yong},
  title = {Towards Clinician-Preferred Segmentation: Leveraging Human-in-the-Loop for Test Time Adaptation in Medical Image Segmentation},
  year = {2024},
  howpublished = {arXiv preprint arXiv:2405.08270},
  doi = {10.48550/arXiv.2405.08270},
  url = {https://arxiv.org/abs/2405.08270}
}

@inproceedings{zhu2025segagent,
  author = {Zhu, Muzhi and Tian, Yuzhuo and Chen, Hao and Zhou, Chunluan and Guo, Qingpei and Liu, Yang and Yang, Ming and Shen, Chunhua},
  title = {{SegAgent}: Exploring Pixel Understanding Capabilities in {MLLMs} by Imitating Human Annotator Trajectories},
  booktitle = {Proceedings of the IEEE/CVF Conference on Computer Vision and Pattern Recognition},
  pages = {3686--3696},
  year = {2025},
  url = {https://openaccess.thecvf.com/content/CVPR2025/html/Zhu_SegAgent_Exploring_Pixel_Understanding_Capabilities_in_MLLMs_by_Imitating_Human_CVPR_2025_paper.html}
}

@article{amershi2014power,
  author  = {Amershi, Saleema and Cakmak, Maya and Knox, W. Bradley and Kulesza, Todd},
  title   = {Power to the People: The Role of Humans in Interactive Machine Learning},
  journal = {AI Magazine},
  volume  = {35},
  number  = {4},
  pages   = {105--120},
  year    = {2014},
  doi     = {10.1609/aimag.v35i4.2513}
}

@inproceedings{antonov2024rclicks,
  author = {Antonov, Anton and Moskalenko, Andrey and Shepelev, Denis and Krapukhin, Alexander and Soshin, Konstantin and Konushin, Anton and Shakhuro, Vlad},
  title = {{RClicks}: Realistic Click Simulation for Benchmarking Interactive Segmentation},
  booktitle = {Advances in Neural Information Processing Systems}, year = {2024}, volume = {37},
  url = {https://proceedings.neurips.cc/paper_files/paper/2024/hash/e6b5bcc872666d37c469e5c5ba723669-Abstract-Datasets_and_Benchmarks_Track.html}
}

@article{archit2025microscopy,
  author = {Archit, Anwai and Freckmann, Luca and Nair, Sushmita and Khalid, Nabeel and Hilt, Paul and Rajashekar, Vikas and Freitag, Marei and Teuber, Carolin and Spitzner, Melanie and Tapia Contreras, Constanza and Buckley, Genevieve and von Haaren, Sebastian and Gupta, Sagnik and Grade, Marian and Wirth, Matthias and Schneider, G{\"u}nter and Dengel, Andreas and Ahmed, Sheraz and Pape, Constantin},
  title = {Segment Anything for Microscopy},
  journal = {Nature Methods}, year = {2025}, volume = {22}, number = {3}, pages = {579--591},
  doi = {10.1038/s41592-024-02580-4}
}

@article{arganda2017weka,
  author = {Arganda-Carreras, Ignacio and Kaynig, Verena and Rueden, Curtis and Eliceiri, Kevin W. and Schindelin, Johannes and Cardona, Albert and Seung, H. Sebastian},
  title = {Trainable {Weka} Segmentation: A Machine Learning Tool for Microscopy Pixel Classification},
  journal = {Bioinformatics}, year = {2017}, volume = {33}, number = {15}, pages = {2424--2426},
  doi = {10.1093/bioinformatics/btx180}
}

@article{bekker2020positive,
  author = {Bekker, Jessa and Davis, Jesse},
  title = {Learning from Positive and Unlabeled Data: A Survey},
  journal = {Machine Learning},
  year = {2020},
  volume = {109},
  pages = {719--760},
  doi = {10.1007/s10994-020-05877-5}
}

@inproceedings{benenson2019large,
  author    = {Benenson, Rodrigo and Popov, Stefan and Ferrari, Vittorio},
  title     = {Large-Scale Interactive Object Segmentation with Human Annotators},
  booktitle = {Proceedings of the IEEE/CVF Conference on Computer Vision and Pattern Recognition},
  pages     = {11700--11709},
  year      = {2019},
  doi       = {10.1109/CVPR.2019.01197}
}

@inproceedings{bengio2009curriculum,
  author = {Bengio, Yoshua and Louradour, J{\'e}r{\^o}me and Collobert, Ronan and Weston, Jason},
  title = {Curriculum Learning},
  booktitle = {Proceedings of the 26th Annual International Conference on Machine Learning},
  pages = {41--48},
  year = {2009},
  publisher = {ACM},
  doi = {10.1145/1553374.1553380}
}

@article{berg2019ilastik,
  author = {Berg, Stuart and Kutra, Dominik and Kroeger, Thorben and Straehle, Christoph N. and Kausler, Bernhard X. and Haubold, Carsten and Schiegg, Martin and Ales, Janez and Beier, Thorsten and Rudy, Markus and Eren, Kemal and Cervantes, Jaime I. and Xu, Buote and Beuttenmueller, Fynn and Wolny, Adrian and Zhang, Chong and Koethe, Ullrich and Hamprecht, Fred A. and Kreshuk, Anna},
  title = {ilastik: Interactive Machine Learning for (Bio)image Analysis},
  journal = {Nature Methods}, year = {2019}, volume = {16}, pages = {1226--1232},
  doi = {10.1038/s41592-019-0582-9}
}

@inproceedings{blier2018description,
  author        = {Blier, L{\'e}onard and Ollivier, Yann},
  title         = {The Description Length of Deep Learning Models},
  booktitle     = {Advances in Neural Information Processing Systems},
  volume        = {31},
  year          = {2018},
  publisher     = {Curran Associates, Inc.},
  eprint        = {1802.07044},
  archivePrefix = {arXiv},
  url           = {https://proceedings.neurips.cc/paper/2018/hash/3b712de48137572f3849aabd5666a4e3-Abstract.html}
}

@inproceedings{boykov2001interactive,
  author    = {Boykov, Yuri Y. and Jolly, Marie-Pierre},
  title     = {Interactive Graph Cuts for Optimal Boundary \& Region Segmentation of Objects in {N-D} Images},
  booktitle = {Proceedings of the Eighth IEEE International Conference on Computer Vision},
  volume    = {1},
  pages     = {105--112},
  year      = {2001},
  doi       = {10.1109/ICCV.2001.937505}
}

@article{cakmak2014eliciting,
  author = {Cakmak, Maya and Thomaz, Andrea L.},
  title = {Eliciting Good Teaching from Humans for Machine Learners},
  journal = {Artificial Intelligence},
  year = {2014},
  volume = {217},
  pages = {198--215},
  doi = {10.1016/j.artint.2014.08.005}
}

@article{cohn1996active,
  author  = {Cohn, David A. and Ghahramani, Zoubin and Jordan, Michael I.},
  title   = {Active Learning with Statistical Models},
  journal = {Journal of Artificial Intelligence Research},
  volume  = {4},
  pages   = {129--145},
  year    = {1996},
  doi     = {10.1613/jair.295},
  url     = {https://www.jair.org/index.php/jair/article/view/10158}
}

@article{dawid1984prequential,
  author = {Dawid, A. P.},
  title = {Present Position and Potential Developments: Some Personal Views. Statistical Theory: The Prequential Approach},
  journal = {Journal of the Royal Statistical Society. Series A (General)},
  volume = {147},
  number = {2},
  pages = {278--292},
  year = {1984},
  doi = {10.2307/2981683}
}

@article{delange2022continual,
  author = {{De Lange}, Matthias and Aljundi, Rahaf and Masana, Marc and Parisot, Sarah and Jia, Xu and Leonardis, Ale\v{s} and Slabaugh, Gregory and Tuytelaars, Tinne},
  title = {A Continual Learning Survey: Defying Forgetting in Classification Tasks},
  journal = {IEEE Transactions on Pattern Analysis and Machine Intelligence},
  year = {2022},
  volume = {44},
  number = {7},
  pages = {3366--3385},
  doi = {10.1109/TPAMI.2021.3057446}
}

@article{diazpinto2024monai,
  author = {Diaz-Pinto, Andres and Alle, Sachidanand and Nath, Vishwesh and Tang, Yucheng and Ihsani, Alvin and Asad, Muhammad and P{\'e}rez-Garc{\'i}a, Fernando and Mehta, Pritesh and Li, Wenqi and Flores, Mona and Roth, Holger R. and Vercauteren, Tom and Xu, Daguang and Dogra, Prerna and Ourselin, Sebastien and Feng, Andrew and Cardoso, M. Jorge},
  title = {{MONAI Label}: A Framework for {AI}-assisted Interactive Labeling of {3D} Medical Images},
  journal = {Medical Image Analysis}, year = {2024}, volume = {95}, pages = {103207},
  doi = {10.1016/j.media.2024.103207}
}

@article{dudley2018review,
  author = {Dudley, John J. and Kristensson, Per Ola},
  title = {A Review of User Interface Design for Interactive Machine Learning},
  journal = {ACM Transactions on Interactive Intelligent Systems},
  year = {2018},
  volume = {8},
  number = {2},
  pages = {8:1--8:37},
  doi = {10.1145/3185517}
}

@article{equitz1991successive,
  author  = {Equitz, William H. R. and Cover, Thomas M.},
  title   = {Successive Refinement of Information},
  journal = {IEEE Transactions on Information Theory},
  volume  = {37},
  number  = {2},
  pages   = {269--275},
  year    = {1991},
  doi     = {10.1109/18.75242}
}

@inproceedings{fails2003interactive,
  author = {Fails, Jerry Alan and Olsen, Jr., Dan R.},
  title = {Interactive Machine Learning},
  booktitle = {Proceedings of the 8th International Conference on Intelligent User Interfaces},
  year = {2003}, pages = {39--45}, doi = {10.1145/604045.604056}
}

@inproceedings{farquhar2021bias,
  author = {Farquhar, Sebastian and Gal, Yarin and Rainforth, Tom},
  title = {On Statistical Bias in Active Learning: How and When to Fix It},
  booktitle = {International Conference on Learning Representations},
  year = {2021},
  url = {https://openreview.net/forum?id=JiYq3eqTKY}
}

@inproceedings{fiebrink2011evaluation,
  author = {Fiebrink, Rebecca and Cook, Perry R. and Trueman, Daniel},
  title = {Human Model Evaluation in Interactive Supervised Learning},
  booktitle = {Proceedings of the SIGCHI Conference on Human Factors in Computing Systems},
  year = {2011},
  pages = {147--156},
  publisher = {ACM},
  doi = {10.1145/1978942.1978965}
}

@article{gneiting2007proper,
  author = {Gneiting, Tilmann and Raftery, Adrian E.},
  title = {Strictly Proper Scoring Rules, Prediction, and Estimation},
  journal = {Journal of the American Statistical Association},
  volume = {102},
  number = {477},
  pages = {359--378},
  year = {2007},
  doi = {10.1198/016214506000001437}
}

@inproceedings{golovin2010nearoptimal,
  author    = {Golovin, Daniel and Krause, Andreas and Ray, Debajyoti},
  title     = {Near-Optimal {Bayesian} Active Learning with Noisy Observations},
  booktitle = {Advances in Neural Information Processing Systems},
  volume    = {23},
  pages     = {766--774},
  year      = {2010},
  publisher = {Curran Associates, Inc.},
  url       = {https://proceedings.neurips.cc/paper/2010/hash/1e6e0a04d20f50967c64dac2d639a577-Abstract.html}
}

@article{golovin2011adaptive,
  author  = {Golovin, Daniel and Krause, Andreas},
  title   = {Adaptive Submodularity: Theory and Applications in Active Learning and Stochastic Optimization},
  journal = {Journal of Artificial Intelligence Research},
  volume  = {42},
  pages   = {427--486},
  year    = {2011},
  doi     = {10.1613/jair.3278},
  url     = {https://arxiv.org/abs/1003.3967v5},
  note    = {Consulted with the corrected arXiv revision of 6 December 2017}
}

@book{grunwald2007mdl,
  author    = {Gr{\"u}nwald, Peter D.},
  title     = {The Minimum Description Length Principle},
  publisher = {MIT Press},
  year      = {2007},
  doi       = {10.7551/mitpress/4643.001.0001}
}

@article{grunwald2020revisited,
  author  = {Gr{\"u}nwald, Peter and Roos, Teemu},
  title   = {Minimum Description Length Revisited},
  journal = {International Journal of Mathematics for Industry},
  volume  = {11},
  number  = {1},
  pages   = {1930001},
  year    = {2020},
  doi     = {10.1142/S2661335219300018}
}

@inproceedings{haertel2015roi,
  author = {Haertel, Robbie and Ringger, Eric K. and Felt, Paul and Seppi, Kevin},
  title = {An Analytic and Empirical Evaluation of Return-on-Investment-Based Active Learning},
  booktitle = {Proceedings of the 9th Linguistic Annotation Workshop},
  year = {2015},
  pages = {11--20},
  publisher = {Association for Computational Linguistics},
  doi = {10.3115/v1/W15-1602},
  url = {https://aclanthology.org/W15-1602/}
}

@article{houlsby2011bald,
  author  = {Houlsby, Neil and Husz{\'a}r, Ferenc and Ghahramani, Zoubin and Lengyel, M{\'a}t{\'e}},
  title   = {Bayesian Active Learning for Classification and Preference Learning},
  journal = {arXiv preprint arXiv:1112.5745},
  year    = {2011},
  doi     = {10.48550/arXiv.1112.5745},
  url     = {https://arxiv.org/abs/1112.5745}
}

@inproceedings{houlsby2012collaborative,
  author    = {Houlsby, Neil and Hern{\'a}ndez-Lobato, Jos{\'e} Miguel and Husz{\'a}r, Ferenc and Ghahramani, Zoubin},
  title     = {Collaborative {Gaussian} Processes for Preference Learning},
  booktitle = {Advances in Neural Information Processing Systems},
  volume    = {25},
  year      = {2012},
  publisher = {Curran Associates, Inc.},
  url       = {https://proceedings.neurips.cc/paper_files/paper/2012/hash/afdec7005cc9f14302cd0474fd0f3c96-Abstract.html}
}

@article{howard2021confidence,
  author = {Howard, Steven R. and Ramdas, Aaditya and McAuliffe, Jon and Sekhon, Jasjeet},
  title = {Time-uniform, Nonparametric, Nonasymptotic Confidence Sequences},
  journal = {The Annals of Statistics},
  volume = {49},
  number = {2},
  pages = {1055--1080},
  year = {2021},
  doi = {10.1214/20-AOS1991}
}

@inproceedings{kendall2017uncertainties,
  author    = {Kendall, Alex and Gal, Yarin},
  title     = {What Uncertainties Do We Need in {Bayesian} Deep Learning for Computer Vision?},
  booktitle = {Advances in Neural Information Processing Systems},
  volume    = {30},
  year      = {2017},
  url       = {https://proceedings.neurips.cc/paper_files/paper/2017/hash/2650d6089a6d640c5e85b2b88265dc2b-Abstract.html}
}

@inproceedings{kirillov2023sam,
  author = {Kirillov, Alexander and Mintun, Eric and Ravi, Nikhila and Mao, Hanzi and Rolland, Chloe and Gustafson, Laura and Xiao, Tete and Whitehead, Spencer and Berg, Alexander C. and Lo, Wan-Yen and Doll{\'a}r, Piotr and Girshick, Ross},
  title = {Segment Anything},
  booktitle = {Proceedings of the IEEE/CVF International Conference on Computer Vision},
  year = {2023}, doi = {10.1109/ICCV51070.2023.00371}
}

@inproceedings{kirsch2019batchbald,
  author    = {Kirsch, Andreas and van Amersfoort, Joost and Gal, Yarin},
  title     = {{BatchBALD}: Efficient and Diverse Batch Acquisition for Deep {Bayesian} Active Learning},
  booktitle = {Advances in Neural Information Processing Systems},
  volume    = {32},
  pages     = {7026--7037},
  year      = {2019},
  publisher = {Curran Associates, Inc.},
  url       = {https://proceedings.neurips.cc/paper_files/paper/2019/hash/95323660ed2124450caaac2c46b5ed90-Abstract.html}
}

@inproceedings{kohl2018probabilistic,
  author    = {Kohl, Simon and Romera-Paredes, Bernardino and Meyer, Clemens and De Fauw, Jeffrey and Ledsam, Joseph R. and Maier-Hein, Klaus and Eslami, S. M. Ali and Jimenez Rezende, Danilo and Ronneberger, Olaf},
  title     = {A Probabilistic {U-Net} for Segmentation of Ambiguous Images},
  booktitle = {Advances in Neural Information Processing Systems},
  volume    = {31},
  pages     = {6965--6975},
  year      = {2018},
  publisher = {Curran Associates, Inc.},
  url       = {https://proceedings.neurips.cc/paper_files/paper/2018/hash/473447ac58e1cd7e96172575f48dca3b-Abstract.html}
}

@inproceedings{lakkaraju2017selective,
  author    = {Lakkaraju, Himabindu and Kleinberg, Jon and Leskovec, Jure and Ludwig, Jens and Mullainathan, Sendhil},
  title     = {The Selective Labels Problem: Evaluating Algorithmic Predictions in the Presence of Unobservables},
  booktitle = {Proceedings of the 23rd ACM SIGKDD International Conference on Knowledge Discovery and Data Mining},
  pages     = {275--284},
  year      = {2017},
  doi       = {10.1145/3097983.3098066}
}

@inproceedings{liu2017iterative,
  author    = {Liu, Weiyang and Dai, Bo and Humayun, Ahmad and Tay, Charlene and Yu, Chen and Smith, Linda B. and Rehg, James M. and Song, Le},
  title     = {Iterative Machine Teaching},
  booktitle = {Proceedings of the 34th International Conference on Machine Learning},
  series    = {Proceedings of Machine Learning Research},
  volume    = {70},
  pages     = {2149--2158},
  year      = {2017},
  url       = {https://proceedings.mlr.press/v70/liu17b.html}
}

@inproceedings{liu2018blackbox,
  author    = {Liu, Weiyang and Dai, Bo and Li, Xingguo and Liu, Zhen and Rehg, James M. and Song, Le},
  title     = {Towards Black-Box Iterative Machine Teaching},
  booktitle = {Proceedings of the 35th International Conference on Machine Learning},
  series    = {Proceedings of Machine Learning Research},
  volume    = {80},
  pages     = {3141--3149},
  year      = {2018},
  url       = {https://proceedings.mlr.press/v80/liu18b.html}
}

@inproceedings{liu2023simpleclick,
  author    = {Liu, Qin and Xu, Zhenlin and Bertasius, Gedas and Niethammer, Marc},
  title     = {{SimpleClick}: Interactive Image Segmentation with Simple Vision Transformers},
  booktitle = {Proceedings of the IEEE/CVF International Conference on Computer Vision},
  pages     = {22290--22300},
  year      = {2023},
  doi       = {10.1109/ICCV51070.2023.02037}
}

@inproceedings{macglashan2017interactive,
  author    = {MacGlashan, James and Ho, Mark K. and Loftin, Robert and Peng, Bei and Wang, Guan and Roberts, David L. and Taylor, Matthew E. and Littman, Michael L.},
  title     = {Interactive Learning from Policy-Dependent Human Feedback},
  booktitle = {Proceedings of the 34th International Conference on Machine Learning},
  series    = {Proceedings of Machine Learning Research},
  volume    = {70},
  pages     = {2285--2294},
  year      = {2017},
  url       = {https://proceedings.mlr.press/v70/macglashan17a.html}
}

@inproceedings{mackowiak2018cereals,
  author    = {Mackowiak, Radek and Lenz, Philip and Ghori, Omair and Diego, Ferran and Lange, Oliver and Rother, Carsten},
  title     = {{CEREALS}: Cost-Effective Region-Based Active Learning for Semantic Segmentation},
  booktitle = {British Machine Vision Conference},
  year      = {2018},
  url       = {https://bmva-archive.org.uk/bmvc/2018/contents/papers/0437.pdf}
}

@article{marinov2024review,
  author = {Marinov, Zdravko and J{\"a}ger, Paul F. and Egger, Jan and Kleesiek, Jens and Stiefelhagen, Rainer},
  title = {Deep Interactive Segmentation of Medical Images: A Systematic Review and Taxonomy},
  journal = {IEEE Transactions on Pattern Analysis and Machine Intelligence},
  year = {2024}, volume = {46}, number = {12}, pages = {10998--11018},
  doi = {10.1109/TPAMI.2024.3452629}
}

@inproceedings{monteiro2020stochastic,
  author    = {Monteiro, Miguel and Le Folgoc, Loic and Coelho de Castro, Daniel and Pawlowski, Nick and Marques, Bernardo and Kamnitsas, Konstantinos and van der Wilk, Mark and Glocker, Ben},
  title     = {Stochastic Segmentation Networks: Modelling Spatially Correlated Aleatoric Uncertainty},
  booktitle = {Advances in Neural Information Processing Systems},
  volume    = {33},
  pages     = {12756--12767},
  year      = {2020},
  publisher = {Curran Associates, Inc.},
  url       = {https://proceedings.neurips.cc/paper_files/paper/2020/hash/95f8d9901ca8878e291552f001f67692-Abstract.html}
}

@inproceedings{nokleby2015ratedistortion,
  author    = {Nokleby, Matthew and Beirami, Ahmad and Calderbank, Robert},
  title     = {A Rate-Distortion Framework for Supervised Learning},
  booktitle = {IEEE 25th International Workshop on Machine Learning for Signal Processing},
  pages     = {1--6},
  year      = {2015},
  doi       = {10.1109/MLSP.2015.7324319}
}

@article{pachitariu2022cellpose,
  author = {Pachitariu, Marius and Stringer, Carsen},
  title = {Cellpose 2.0: How to Train Your Own Model},
  journal = {Nature Methods}, year = {2022}, volume = {19}, pages = {1634--1641},
  doi = {10.1038/s41592-022-01663-4}
}

@inproceedings{perdomo2020performative,
  author    = {Perdomo, Juan and Zrnic, Tijana and Mendler-D{\"u}nner, Celestine and Hardt, Moritz},
  title     = {Performative Prediction},
  booktitle = {Proceedings of the 37th International Conference on Machine Learning},
  series    = {Proceedings of Machine Learning Research},
  volume    = {119},
  pages     = {7599--7609},
  year      = {2020},
  url       = {https://proceedings.mlr.press/v119/perdomo20a.html}
}

@article{ren2021survey,
  author = {Ren, Pengzhen and Xiao, Yun and Chang, Xiaojun and Huang, Po-Yao and Li, Zhihui and Gupta, Brij B. and Chen, Xiaojiang and Wang, Xin},
  title = {A Survey of Deep Active Learning},
  journal = {ACM Computing Surveys},
  year = {2021},
  volume = {54},
  number = {9},
  pages = {180:1--180:40},
  doi = {10.1145/3472291}
}

@article{rissanen1978modeling,
  author  = {Rissanen, Jorma},
  title   = {Modeling by Shortest Data Description},
  journal = {Automatica},
  volume  = {14},
  number  = {5},
  pages   = {465--471},
  year    = {1978},
  doi     = {10.1016/0005-1098(78)90005-5}
}

@article{rother2004grabcut,
  author  = {Rother, Carsten and Kolmogorov, Vladimir and Blake, Andrew},
  title   = {{GrabCut}: Interactive Foreground Extraction Using Iterated Graph Cuts},
  journal = {ACM Transactions on Graphics},
  volume  = {23},
  number  = {3},
  pages   = {309--314},
  year    = {2004},
  doi     = {10.1145/1015706.1015720}
}

@article{rubin1976missing,
  author = {Rubin, Donald B.},
  title = {Inference and Missing Data},
  journal = {Biometrika},
  year = {1976},
  volume = {63},
  number = {3},
  pages = {581--592},
  doi = {10.1093/biomet/63.3.581}
}

@inproceedings{sener2018active,
  author    = {Sener, Ozan and Savarese, Silvio},
  title     = {Active Learning for Convolutional Neural Networks: A Core-Set Approach},
  booktitle = {International Conference on Learning Representations},
  year      = {2018},
  url       = {https://openreview.net/forum?id=H1aIuk-RW}
}

@techreport{settles2009active,
  author      = {Settles, Burr},
  title       = {Active Learning Literature Survey},
  institution = {University of Wisconsin--Madison},
  number      = {1648},
  year        = {2009},
  url         = {https://burrsettles.com/pub/settles.activelearning_20090109.pdf}
}

@article{shafto2014rational,
  author = {Shafto, Patrick and Goodman, Noah D. and Griffiths, Thomas L.},
  title = {A Rational Account of Pedagogical Reasoning: Teaching by, and Learning from, Examples},
  journal = {Cognitive Psychology},
  year = {2014},
  volume = {71},
  pages = {55--89},
  doi = {10.1016/j.cogpsych.2013.12.004}
}

@article{shivaswamy2015coactive,
  author  = {Shivaswamy, Pannaga and Joachims, Thorsten},
  title   = {Coactive Learning},
  journal = {Journal of Artificial Intelligence Research},
  volume  = {53},
  pages   = {1--40},
  year    = {2015},
  doi     = {10.1613/jair.4539}
}

@article{smith2022rootpainter,
  author  = {Smith, Abraham George and Han, Eusun and Petersen, Jens and Olsen, Niels Alvin Faircloth and Giese, Christian and Athmann, Miriam and Dresb\o{}ll, Dorte Bodin and Thorup-Kristensen, Kristian},
  title   = {{RootPainter}: Deep Learning Segmentation of Biological Images with Corrective Annotation},
  journal = {New Phytologist},
  volume  = {236},
  number  = {2},
  pages   = {774--791},
  year    = {2022},
  doi     = {10.1111/nph.18387}
}

@article{smith2022rootpainter3d,
  author  = {Smith, Abraham George and Petersen, Jens and Terrones-Campos, Cynthia and Berthelsen, Anne Kiil and Forbes, Nora Jarrett and Darkner, Sune and Specht, Lena and Vogelius, Ivan Richter},
  title   = {{RootPainter3D}: Interactive-Machine-Learning Enables Rapid and Accurate Contouring for Radiotherapy},
  journal = {Medical Physics},
  volume  = {49},
  number  = {1},
  pages   = {461--473},
  year    = {2022},
  doi     = {10.1002/mp.15353}
}

@inproceedings{sofiiuk2022focalclick,
  author    = {Chen, Xi and Zhao, Zhiyan and Zhang, Yilei and Duan, Manni and Qi, Donglian and Zhao, Hengshuang},
  title     = {{FocalClick}: Towards Practical Interactive Image Segmentation},
  booktitle = {Proceedings of the IEEE/CVF Conference on Computer Vision and Pattern Recognition},
  pages     = {1300--1309},
  year      = {2022},
  doi       = {10.1109/CVPR52688.2022.00136}
}

@inproceedings{top2011active,
  author    = {Top, Andrew and Hamarneh, Ghassan and Abugharbieh, Rafeef},
  title     = {Active Learning for Interactive {3D} Image Segmentation},
  booktitle = {Medical Image Computing and Computer-Assisted Intervention -- {MICCAI} 2011},
  series    = {Lecture Notes in Computer Science},
  volume    = {6893},
  pages     = {603--610},
  year      = {2011},
  doi       = {10.1007/978-3-642-23626-6_74}
}

@inproceedings{vezhnevets2012expected,
  author    = {Vezhnevets, Alexander and Buhmann, Joachim M. and Ferrari, Vittorio},
  title     = {Active Learning for Semantic Segmentation with Expected Change},
  booktitle = {Proceedings of the IEEE Conference on Computer Vision and Pattern Recognition},
  pages     = {3162--3169},
  year      = {2012},
  doi       = {10.1109/CVPR.2012.6248050}
}

@article{wang2018bifseg,
  author = {Wang, Guotai and Li, Wenqi and Zuluaga, Maria A. and Pratt, Rosalind and Patel, Premal A. and Aertsen, Michael and Doel, Tom and David, Anna L. and Deprest, Jan and Ourselin, S{\'e}bastien and Vercauteren, Tom},
  title = {Interactive Medical Image Segmentation Using Deep Learning With Image-Specific Fine Tuning},
  journal = {IEEE Transactions on Medical Imaging}, year = {2018}, volume = {37}, number = {7}, pages = {1562--1573},
  doi = {10.1109/TMI.2018.2791721}
}

@article{wang2020errortolerant,
  author  = {Wang, Tao and Qi, Shengzhe and Jia, Zexuan and Sun, Quansen and Fu, Peng and Ge, Qi},
  title   = {Error-Tolerant Label Prior for Interactive Image Segmentation},
  journal = {Information Sciences},
  volume  = {538},
  pages   = {384--395},
  year    = {2020},
  doi     = {10.1016/j.ins.2020.05.122}
}

@article{wu2022human,
  author = {Wu, Xingjiao and Xiao, Luwei and Sun, Yixuan and Zhang, Junhang and Ma, Tianlong and He, Liang},
  title = {A Survey of Human-in-the-Loop for Machine Learning},
  journal = {Future Generation Computer Systems},
  year = {2022},
  volume = {135},
  pages = {364--381},
  doi = {10.1016/j.future.2022.05.014}
}

@inproceedings{yan2011crowds,
  author    = {Yan, Yan and Rosales, R{\'o}mer and Fung, Glenn and Dy, Jennifer G.},
  title     = {Active Learning from Crowds},
  booktitle = {Proceedings of the 28th International Conference on Machine Learning},
  year      = {2011},
  url       = {https://icml.cc/2011/papers/596_icmlpaper.pdf}
}

@inproceedings{yuan2021teacheraware,
  author = {Yuan, Luyao and Zhou, Dongruo and Shen, Juhong and Gao, Jingdong and Chen, Jeffrey L. and Gu, Quanquan and Wu, Ying Nian and Zhu, Song-Chun},
  title = {Iterative Teacher-Aware Learning},
  booktitle = {Advances in Neural Information Processing Systems},
  volume = {34},
  year = {2021},
  url = {https://proceedings.neurips.cc/paper/2021/hash/f48c04ffab49ff0e5d1176244fdfb65c-Abstract.html}
}

@article{yuan2023communicative,
  author  = {Yuan, Luyao and Zhu, Song-Chun},
  title   = {Communicative Learning: A Unified Learning Formalism},
  journal = {Engineering},
  volume  = {25},
  number  = {6},
  pages   = {77--100},
  month   = jun,
  year    = {2023},
  doi     = {10.1016/j.eng.2022.10.017}
}

@article{zhang2023multiple,
  author  = {Zhang, Le and Tanno, Ryutaro and Xu, Moucheng and Huang, Yawen and Bronik, Kevin and Jin, Chen and Jacob, Joseph and Zheng, Yefeng and Shao, Ling and Ciccarelli, Olga and Barkhof, Frederik and Alexander, Daniel C.},
  title   = {Learning from Multiple Annotators for Medical Image Segmentation},
  journal = {Pattern Recognition},
  volume  = {138},
  pages   = {109400},
  year    = {2023},
  doi     = {10.1016/j.patcog.2023.109400}
}

@article{jaiswal2026xpertloop,
  author = {Jaiswal, Astha and Rinneburger, Miriam and Meyer, Franziska and Arjune, Sita and Oberlinkels, Laura and Wawer Matos Reimer, Philomena Alice and Lotter-Becker, Lars and Ak{\"u}nal, {\"U}nal and Bujotzek, Markus and Denner, Stefan and Maier-Hein, Klaus and Stepansky, Leonard and May, Matthias Stefan and Habert, Markus and K{\"o}hn, Alexander and Sch{\"o}neck, Mirjam and Wawer Matos Reimer, Robert Peter and M{\"u}ller, Roman-Ulrich and Lennartz, Simon and Gro{\ss}e Hokamp, Nils and Bucher, Andreas Michael and Persigehl, Thorsten and Caldeira, Liliana Lourenco},
  title = {Impact on Cost and Expert Time of Data-Efficient Deep Learning for Medical Image Segmentation},
  journal = {Radiology: Artificial Intelligence},
  year = {2026},
  volume = {8},
  number = {4},
  pages = {e250200},
  doi = {10.1148/ryai.250200}
}

@article{dendukuri2026active,
  author = {Dendukuri, Manasa and Jogan, Matja{\v z} and Hashimoto, Daniel A. and Liao, Guiqiu},
  title = {Active Learning for Efficient Annotation of Surgical Videos with Weak Supervision},
  journal = {arXiv preprint arXiv:2607.13237},
  year = {2026},
  doi = {10.48550/arXiv.2607.13237}
}

@article{shaeri2025explainable,
  author = {Shaeri, Pouya and Woo, Ryan T. and Mohammadpour, Yasaman and Middel, Ariane},
  title = {Explainable Human-in-the-Loop Segmentation via Critic Feedback Signals},
  journal = {arXiv preprint arXiv:2510.09945},
  year = {2025},
  doi = {10.48550/arXiv.2510.09945}
}

@inproceedings{tavangar2026exil,
  author = {Tavangar, Arash and Chiu, Larissa K. and Khodashenas, Hamidreza and Berry, Gregory K. and Hooshiar, Amir},
  title = {Expert-like Bone Ultrasound Segmentation through Expert-in-the-loop Mask-conditioned Progressive Learning},
  booktitle = {Medical Image Computing and Computer Assisted Intervention -- MICCAI 2026},
  series = {Lecture Notes in Computer Science},
  volume = {16884},
  publisher = {Springer Nature Switzerland},
  year = {2026},
  url = {https://papers.miccai.org/miccai-2026/0361-Paper6332.html}
}

@inproceedings{xu2026youpoint,
  author = {Xu, Wentian and Liang, Ziyun and Anthony, Harry and Ibrahim, Yasin and Cohen, Felix and Yang, Guang and Kamnitsas, Konstantinos},
  title = {You Point, {I} Learn: Online Adaptation of Interactive Segmentation Models for Handling Distribution Shifts in Medical Imaging},
  booktitle = {Proceedings of the International Conference on Learning Representations},
  year = {2026},
  url = {https://proceedings.iclr.cc/paper_files/paper/2026/hash/8e5e1e120afdbd8700ee4ce8c42f2bf5-Abstract-Conference.html}
}

@inproceedings{wong2025multiverseg,
  author = {Wong, Hallee E. and Gonzalez Ortiz, Jose Javier and Guttag, John and Dalca, Adrian V.},
  title = {{MultiverSeg}: Scalable Interactive Segmentation of Biomedical Imaging Datasets with In-Context Guidance},
  booktitle = {Proceedings of the IEEE/CVF International Conference on Computer Vision},
  year = {2025},
  pages = {20966--20980},
  doi = {10.1109/ICCV51701.2025.01949}
}

@inproceedings{torpey2026learning,
  author = {Torpey, Gianna and Guttag, John and Wong, Hallee E.},
  title = {Learning What To Ask For When: Image Ordering for In-Context Interactive Medical Image Segmentation},
  booktitle = {Proceedings of the IEEE/CVF Conference on Computer Vision and Pattern Recognition Workshops},
  year = {2026},
  month = jun,
  pages = {6418--6426},
  url = {https://openaccess.thecvf.com/content/CVPR2026W/PHAROS-AIF-MIH/html/Torpey_Learning_What_To_Ask_For_When_Image_Ordering_for_In-Context_CVPRW_2026_paper.html}
}

@article{zhu2018overview,
  author  = {Zhu, Xiaojin and Singla, Adish and Zilles, Sandra and Rafferty, Anna N.},
  title   = {An Overview of Machine Teaching},
  journal = {arXiv preprint arXiv:1801.05927},
  year    = {2018},
  url     = {https://arxiv.org/abs/1801.05927}
}
\end{document}